\documentclass[runningheads]{llncs}
\usepackage{graphicx}
\graphicspath{{./figures/}}

\usepackage{tikz}
\usepackage{comment}
\usepackage{amsmath,amssymb} % define this before the line numbering.
\usepackage{color}

\usepackage[width=122mm,left=12mm,paperwidth=146mm,height=193mm,top=12mm,paperheight=217mm]{geometry}

\usepackage{subfigure}
\usepackage[hidelinks]{hyperref}
\usepackage{tabularx}
\usepackage{makecell,multirow,diagbox}

\usepackage{bbding}

\usepackage[marginal]{footmisc}

\newcommand{\dbname}[1]{\emph{#1}}

\begin{document}
% \linenumbers
\pagestyle{headings}
\mainmatter

\title{A Joint Convolution Auto-encoder Network for Infrared and Visible Image Fusion} % Replace with your title

%\begin{comment}
%\titlerunning{BICS2020 submission} 
%\authorrunning{BICS2020 submission} 
%\author{Xiaoqinig Luo}
%\institute{Jiangnan University}
%\end{comment}
%******************

%\begin{comment}
\titlerunning{JCAE for Infrared and Visible Image Fusion}

\author{Zhancheng Zhang\inst{1} \and
Yuanhao Gao\inst{2} \and
Mengyu Xiong\inst{2} \and
Xiaoqing Luo\inst{2(}\Envelope\inst{)} \and
Xiao-Jun Wu\inst{2}}
\authorrunning{X. Luo et al.}
\institute{Suzhou University of Science and Technology, Suzhou, China \and
Jiangnan University, Wuxi, China\\
\email{xqluo@jiangnan.edu.cn}
}
%\end{comment}
%******************
\maketitle

%\footnote{Corresponding  author: Xiaoqing Luo, xqluo@jiangnan.edu.cn\\
%Conflict of Interest:  Xiaoqing Luo has received research grants from the National Natural Science Foundation of P. R. China under grant no. 61772237 and the Six Talent Peaks Project in Jiangsu Province under grant XYDXX-030. Yuanhao Gao, Mengyu Xiong, Zhancheng Zhang and Xiaojun Wu declare that they have no conflict of interest.}

\begin{abstract}
Background: Leaning redundant and complementary relationships is a critical step in the human visual system. Inspired by the infrared cognition ability of crotalinae animals, we design a joint convolution auto-encoder (JCAE) network for infrared and visible image fusion. Methods: Our key insight is to feed infrared and visible pair images into the network simultaneously and separate an encoder stream into two private branches and one common branch, the private branch works for complementary features learning and the common branch does for redundant features learning. We also build two fusion rules to integrate redundant and complementary features into their fused feature which are then fed into the decoder layer to produce the final fused image. We detail the structure, fusion rule and explain its multi-task loss function. Results: Our JCAE network achieves good results in terms of both subjective effect and objective evaluation metrics.

\end{abstract}

\section{Introduction}
Infrared sensor and visible sensor have different imaging characteristics. Infrared sensor performs well on thermal target but with a blurry background, on the contrary, visible light sensor has more imaging spectrum but in which interesting targets are varied due to the varying light, even these targets might be faint or lost in low light condition \cite{Chen2016Image}. Infrared and visible image fusion can combine infrared targets and more clear background scene into an integrated image which has richer information. Some crotalinae animals have evolved to have the ability of fusion infrared spectrum for hunting, such as snake can exactly catch rat at night. Exploring and simulating this cognition will benefit to image fusion technology.

A lot of infrared and visible image fusion methods have been proposed in the past decades \cite{Chen2009Research,Wang2011A,Chuan2008Fusion,Liu2017Multi,luo2016novel,Luo2021IFSR}. Generally speaking, image fusion methods can be divided into two categories: spatial domain based fusion methods and transform domain based fusion methods. Spatial domain based fusion methods \cite{Chen2009Research,Wang2011A,Liu2017Multi} usually grid-partition whole source images into many small overlapped blocks, each pair of blocks are fused with designed activity measurement. These methods usually assume that candidate images come from the same modal. However, infrared image and visible image are individually captured from sensors with different modal and have the different characteristic. The spatial domain-based methods ignore the modal difference between a pair of source images, and the fusion of pixel blocks has a large difference in the fusion results in the scene with different brightness levels.

The transform domain based methods usually decompose source images into sub-images with different frequency bands by a multi-scale decomposition method in the first step. Then fusion rules are designed according to the characteristics of decomposed band coefficients. Finally, the fused images are obtained through the corresponding inverse transformation. During the whole fusion process, the selection of multi-scale decomposition methods and the design of fusion rules are two important factors affecting fusion quality, among them, many decomposition methods were employed, such as Laplacian pyramid \cite{Wang2011A}, pyramid decomposition  \cite{Chen2009Research}, wavelet transform \cite{Pang2012Multifocus}, shearlet transform \cite{luo2017image}, curvelet transform \cite{Qi2009Image} etc. Choose-max fusion strategy and weighted average fusion strategy are two widely used strategies. These methods fuse images from the feature level, but the manually designed rule cannot adaptively produce good results due to the variety of imaging scenery.

In recent years, convolution neural network (CNN) has achieved the state-of-the-art results in many computer vision tasks, such as image super-resolution \cite{Dong2016Accelerating}, target tracking \cite{Bertinetto2016Fully}, semantic segmentation \cite{Shelhamer2014Fully}, recognition \cite{He2015Deep} etc. CNN can learn many effective features from a large amount of training data. By taking advantage of these features, some CNN-based image fusion methods were recently proposed. Liu et al. \cite{Liu2017Multi} cast a multi-focus image fusion problem into a CNN-based binary classification problem. In its first stage, a classifier was trained on many small image blocks with focused/unfocused label according to the blur level of the image block. Then the focus map was predicted by the trained classifier and the result image was fused by a spatial domain based selection rule based on this map. Predicting label map in infrared and visible image fusion tasks is difficult, the reason lies in that infrared and visible images are with different modal and their features vary greatly among modal and image scene, hence, Liu's method cannot be straightly used to fuse infrared and visible images. To overcome the modality gap and the lack of physical ground truth image, Luo et al.\cite{Luo2021Lat} trained the CNN with synthetic infrared and visible image, which can be well generalized to the real-life infrared and visible image fusion tasks. Nevertheless, the deviation of the sythetic images from the real ones can limit the application in some rare cases. Thereby, data-driven NN style infrared and visible image fusion is still an open problem. Li et al. \cite{Hui2018DenseFuse} used an auto-encoder network to combine infrared and visible images to get advanced results, in that work, densenet are used to extract features of an image, which makes the detailed information well-preserved. However, the extracted features are not well distinguished, so building an adaptive rule to fuse these features is still an open problem. 

On unlabeled data, learning features in an unsupervised manner performs well.  Following this style, the auto-encoder network is a wide-used backend network, where feature extraction function is realized by encoder layers and reconstruction function is realized by decoder layers. Moreover, by combining more layers including convolution layers, pooling layers, and up-sampling layers, convolution auto-encoder can learn more robust and hierarchical features. For dealing with multi-exposure image fusion problem, Prabhakar et al. \cite{Prabhakar2017DeepFuse} proposed a method based on convolution neural network which was trained on multiple exposure image pairs themselves. As a data-driven and self-taught fusion method, the network has two convolution layers for encoding and three convolution layers for decoding, particularly, the weights of its encoder layers are shared, the network structure is forced to learn the same features from the input image pair, then the learned features can be simply processed in the fusion layer. Sum fusion strategy was employed to integrate the learned features which were then fed into the decoder layers to reconstruct a well-exposure image. For incorporating prior knowledge into feature learning, Baruch Epstein et al. \cite{Meir2017Joint} proposed an auto-encoder network with a  multi-task parallel structure. This network consists of common branches and private branches, the common branch tends to learn some common features by sharing weights during training, while the private branch does not share weights and tends to learn specific features. For infrared and visible image fusion, although  different modal, the pair of images are captured on the same field of view and intrinsically aligned, thus their redundancy (says the same field of view) and complementarity (says different modal) relationship can be considered as prior knowledge, and to distinguish this relationship is essential in a fusion process. 

Motivated by these findings, we propose a data-driven joint convolution auto-encoder (JCAE) network for fusing infrared and visible images. Firstly, we use multi-branch structure and feature sharing to extract redundant features and complementary features from a pair of source images. After that, we design corresponding fusion rules according to the extracted features, these fused features are then fed into the decoding layer, getting a fused image in an end-to-end manner. For enhancing JCAE's feature learning ability, part layers of VGG19 are transferred into the encoder layers , and an image quality related multi-task loss function is added to the network. More than 1400 pairs of grayscale images of the IFCNN \cite{2020IFCNN} dataset are used as our training set. Finally, 48 pairs of infrared and visible images of the TNO dataset are used as the test set. The major contributions of this paper are as follows:
\begin{enumerate}
	\item Inspired by the infrared cognitive ability of crotalinae animals, propose a joint convolution auto-encoder network which learns the inherent redundancy and complementarity relationship among paired infrared and visible images. 
	\item Propose a feature layer fusion method based on the joint convolution auto-encoder network to fuse complementary and redundant features in an end-to-end fashion.
	\item Introduce image quality factors into the loss function to improve the fusion quality.
	\item Transfer part layers of VGG19 into the joint convolution auto-encoder and enhance the fused image quality. 
\end{enumerate}	

\section{Proposed Method} \label{sec:propsed_method}
In this section, JCAE's joint learning method, network structure and fusion quality oriented multi-task loss function are introduced, then the method of transferring VGG19's into JCAE is discussed. Finally, JCAE-based fusion rule are built to integrate the learned features and reconstruct a fused image.

\subsection{How private branch and common branch jointly work}\label{sec:joint_work}
\begin{figure}[!htbp]
	\addtolength{\abovecaptionskip}{-5pt}
	\addtolength{\belowcaptionskip}{-10pt}
	\centering
	\subfigure[]{	
		\begin{minipage}[c]{0.3\linewidth}
			\centering
			\includegraphics[width=\textwidth]{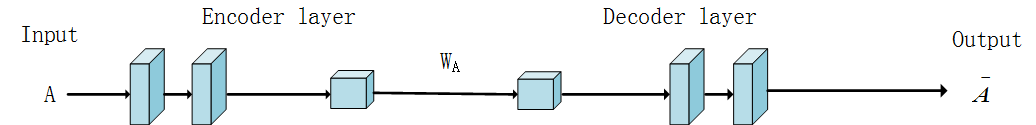}
			%\caption{fig.1}
		\end{minipage}			
	}%	
	\subfigure[]{
		\begin{minipage}[c]{0.3\linewidth}
			\centering
			\includegraphics[width=\textwidth]{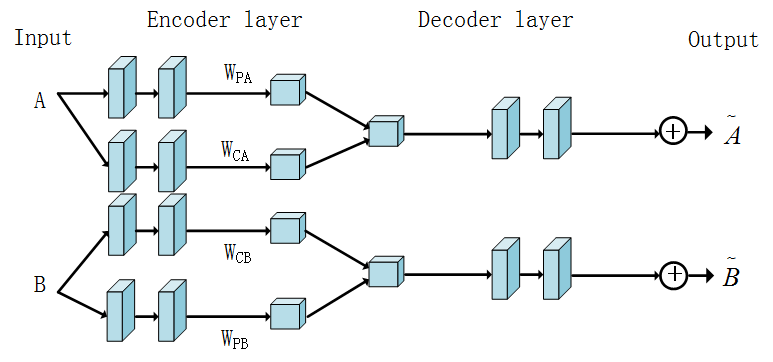}
			%\caption{fig2}		
		\end{minipage}
	}%
	\subfigure[]{
		\begin{minipage}[c]{0.4\linewidth}
			\centering
			\includegraphics[width=\textwidth]{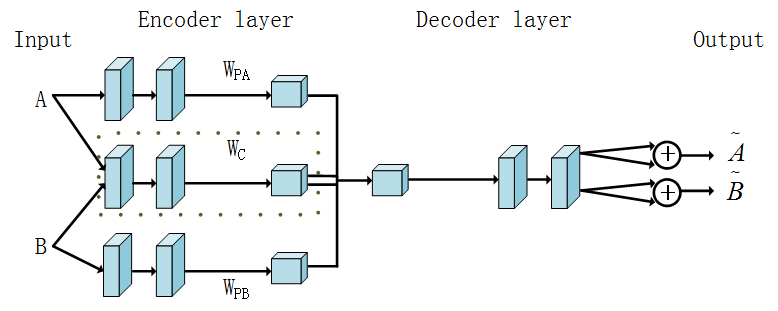}
			%\caption{fig3}				
		\end{minipage}
	}%
	\caption{Auto-encoder network structure. (a) Traditional auto-encoder network; (b) Auto-encoder network for joint training; (c) Joint auto-encoder network owns sharing branches.}
	%\textbf{Fig.1.}~~The proposed fusion method
	\label{fig:autonetwork}
\end{figure}

Auto-encoder network is a type of unsupervised learning network, its hidden layer can be considered as feature representation of input images. Taking image $A$ as an example, its encoding and decoding processes are illustrated as \autoref{fig:autonetwork}(a), this process can be formulated as equation (\ref{eq:1}):
\begin{equation}
\min \left(A - \left({g_{dec}} \circ {g_{enc}}\right)\left(A\right)\right)
\label{eq:1}
\end{equation}
where $g_{enc}$ represents encoding process, $g_{dec}$ does for decoding. Generally, a loss function forces the  reconstructed image to be as close as the original input $A$, hence the learned ${g_{enc}}(A)$ is the latent representation of $A$.

Due to the fact that infrared image $A$ and visible image $B$ are captured on the same visual field and intrinsically aligned in a fusion task. For utilizing this prior knowledge, we try to divide the network of \autoref{fig:autonetwork}(a) into two individual branches: one for learning the common representation from the same visual field, and another for learning the private representation from the different imaging modal. To encoder $A$ and $B$, the network illustrated in \autoref{fig:autonetwork}(a) can be re-built into 
that of \autoref{fig:autonetwork}(b) and equation (\ref{eq:1}) can be rewritten as: 	
\begin{equation}
\left\{ \begin{array}{l}
\min \left(A - \left(\left(g_{dec}^{CA} \circ g_{enc}^{CA}\right)\left(A\right) + \left(g_{dec}^{PA} \circ g_{enc}^{PA}\right)\left(A\right)\right)\right)\\
\min \left(B - \left(\left(g_{dec}^{CB} \circ g_{enc}^{CB}\right)\left(B\right) + \left(g_{dec}^{PB} \circ g_{enc}^{PB}\right)\left(B\right)\right)\right)
\end{array} \right.
\label{eq:2}
\end{equation}
In this manner, image $A$ and $B$ can be reconstructed by combining their common features $(g_{dec}^{CA} \circ g_{enc}^{CA})(A)$ and $(g_{dec}^{CB} \circ g_{enc}^{CB})(B)$ and private features $(g_{dec}^{PA} \circ g_{enc}^{PA})(A)$ and $(g_{dec}^{PB} \circ g_{enc}^{PB})(B)$ separately, however, $A$ and $B$ are encoded in their corresponding feature spaces, which does not fully agree with the prior knowledge, i.e. the two images come from the same visual field and are pre-registered. To fully embedding this prior knowledge into the feature learning process, we further make the two networks share the same common branch and the same decoding layers. Without loss of generality, one can further assume that:
\begin{equation}
\left\{ \begin{array}{l}
g_{enc}^{CA} = g_{enc}^{CB} = g_{enc}^C\\
g_{dec}^{CA} = g_{dec}^{CB} =g_{dec}^{PA} = g_{dec}^{PB} = g_{dec}^C\\
\end{array} \right.
\label{eq:4}
\end{equation}
Thus, the optimization can be rewritten as follows:
\begin{equation}
\begin{aligned}
\min &\left(\left(A - \left(\left(g_{dec}^C \circ g_{enc}^C\right)\left(A\right) + \left(g_{dec}^{C} \circ g_{enc}^{PA}\right)\left(A\right)\right)\right)\right.\\ 
&+ \left.\left(B - \left(\left(g_{dec}^C \circ g_{enc}^C\right)\left(B\right) + \left(g_{dec}^{C} \circ g_{enc}^{PB}\right)\left(B\right)\right)\right)\right) \label{eq:5}
\end{aligned}	
\end{equation}

In this case, sharing decoding layers makes the network jointly learn features hidden in the paired images, sharing common branches force the network to learn a common feature representation,  and individual private branch tends to learn the features which distinguish one image from another for different imaging modal. This hybrid structure was demonstrated in \autoref{fig:autonetwork}(c). Note, the aforementioned equations are discussed on a general auto-encoder, this transformation can be straightly extended to other specified auto-encoder. In this paper, an auto-encoder with convolution layers are built and its details will be discussed in the following subsection \ref{sec:network_structure}.

\subsection{Network structure} \label{sec:network_structure}	
\begin{figure}[!htbp]
    \addtolength{\abovecaptionskip}{-5pt}
    \addtolength{\belowcaptionskip}{-10pt}
	\centering
	\includegraphics[width=\textwidth]{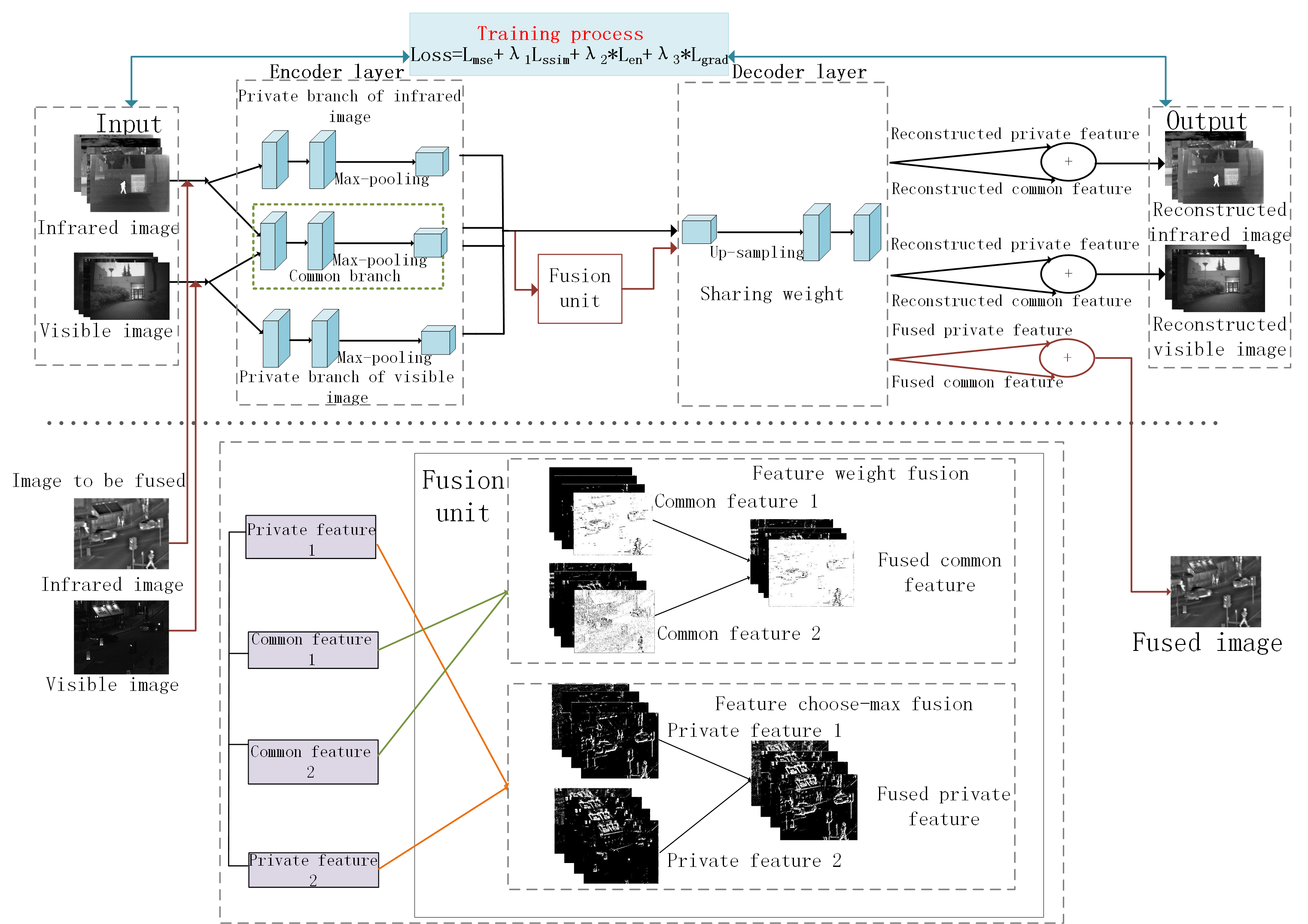}
	\caption{The proposed fusion method.}
	%\textbf{Fig.1.}~~The proposed fusion method
	\label{fig:network}
\end{figure}

When convolution units are combined with the aforementioned joint learning structure, the integrated network is called the joint convolution auto-encoder (JCAE). As shown in \autoref{fig:network}, in an infrared and visible image fusion task, JCAE's encoder layers consists of two private branches and one common branch: the private branches try to learn the features for distinguishing one image from another, these features stands for the  complementary relationship among the paired infrared and visible images, on the other hand, common branch tends to learn some common features which represent the redundant relationship. In the decoder layers, the weights are shared to merge the previous learned private and common features and to reconstruct the input images. The whole fusion process is divided into two stages. The first is the feature learning stage and the second is the fusion stage. 

In the feature learning stage, firstly, we input image pair through the private and common branches of the encoder layer to obtain the private and common features respectively. Training process make the network have the ability of restructuring images through the joint convolution auto-encoder network. The input data and output data of the network are the same data. The encoder layer of the joint convolution auto-encoder network is a typical convolution process, the purpose of the pooling layer is to reduce the dimension of the data and the decoder layer reconstructs the compressed data to the original image. 

In the fusion stage, a pair of infrared and visible images are input to the joint convolution auto-encoder network, the last layer of the encoder layer outputs the common features and private features of the source images respectively. In the training process, common features reflect the redundant relationship of the images while private features reflect the complementary relationship. According to the difference of feature form, we then build the corresponding fusion rule: for private features, they should remain without any loss in the fusion process, thus, we adopt choose-max fusion strategy, the complementary information in the same location of the two source images will remain without any loss. For common features, which usually reflect the general shape and structure information rather than complementary information, so we choose weight fusion strategy and do weighted fusion on these features  in the same location. The fusion process is also shown as a fusion unit in \autoref{fig:network}. By fusing the fused common feature and private feature through the decoder layer, we can get a fused result directly. We will formulate the fusion detail in Section \ref{sec:fusion_rule}. 
	
\subsection{Multi-task loss function}\label{sec:loss_func}	
From the perspective of image reconstruction, using mean squared error (MSE) as loss function in an auto-encoder network can well realize the image reconstruction, however, MSE expresses the overall pixel loss of the image, the edge information of the image can not be well expressed by using MSE alone, and structure of the image may be lost. For making the auto-encoder network better learn the redundancy and complementarity relationship between multi-modal images, we further introduce the structural similarity image metric  (SSIM) \cite{ssim}  into the loss function. The combined loss function can be written as:

\begin{equation}
Loss = {L_{mse}} + {\lambda} \cdot {L_{ssim}} 
\label{eq:6}
\end{equation}
\noindent The MSE loss is formulated as:
\begin{equation}
{L_{mse}} = \sqrt {\frac{1}{{MN}}} \sum\limits_{x = 1}^M {\sum\limits_{y = 1}^N {[I(x,y) - O(x,y)]} }
\label{eq:7}
\end{equation}
where $M$ and $N$ are the size of images. The smaller the mean squared error are, the closer to the original images the reconstructed images are.
\noindent The SSIM loss is formulated as:
\begin{equation}
{L_{ssim}} = 1 - SSIM(O,I)
\label{eq:8}
\end{equation}	
$SSIM(\cdot, \cdot)$ defines the structural similarity of two original images. where $O$ is the output predict image, $I$ is the input image. The setting of parameters in the combined loss function will be discussed in section 3.2.

\subsection{Transferring VGG into JCAE } \label{sec:trans}	
The feature learning ability of VGG19 is straight-forward, the key issue in this paper is which layers are helpful to image fusion. In this subsection, for improving training accuracy and speed of the network, we transfer part layers of VGG19 \cite{Simonyan2014Very} into JCAE.

\begin{figure}[!htbp]
	\addtolength{\abovecaptionskip}{-5pt}
	\addtolength{\belowcaptionskip}{-12pt}
	\def\tempwidth{0.20\textwidth}
	\centering
	\tiny	
	\subfigure[]{\includegraphics[width=\tempwidth]{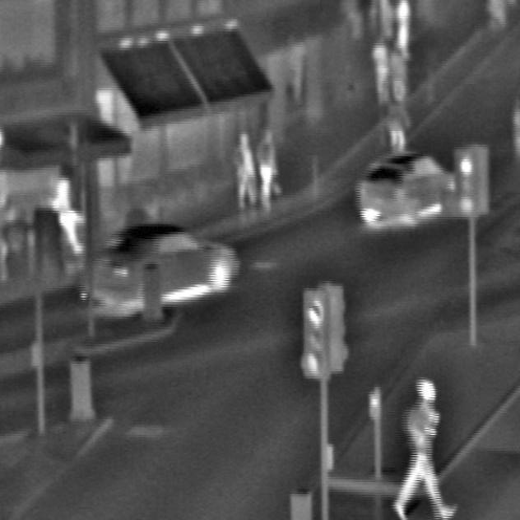}}\,
	\subfigure[]{\includegraphics[width=\tempwidth]{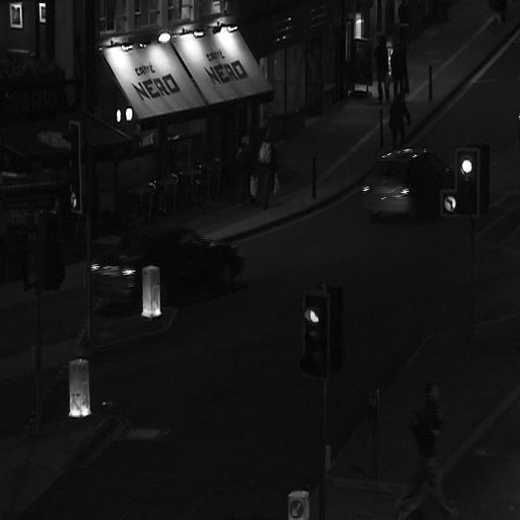}}\,
	\subfigure[]{\includegraphics[width=\tempwidth]{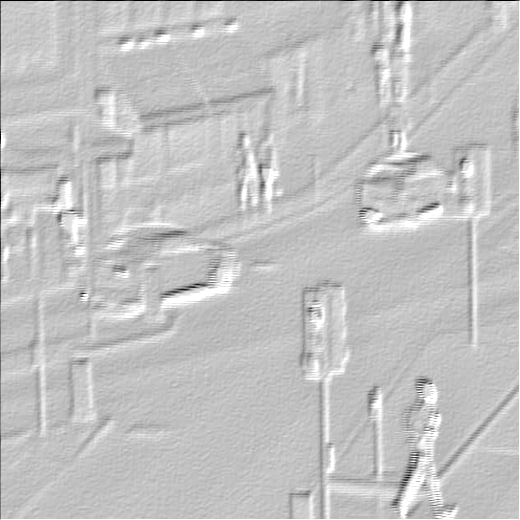}}\,
	\subfigure[]{\includegraphics[width=\tempwidth]{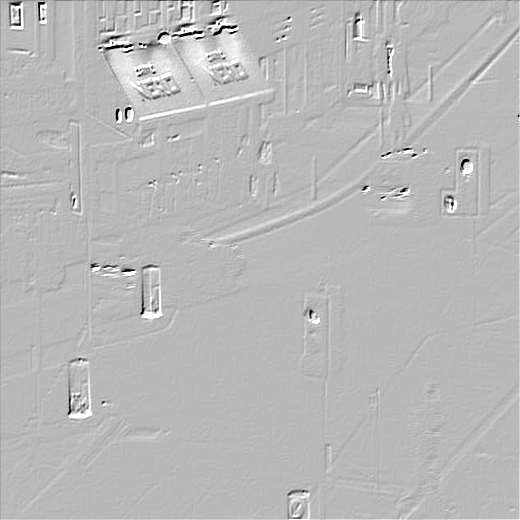}}
	\\
	\vspace{-2mm}
	\subfigure[]{\includegraphics[width=\tempwidth]{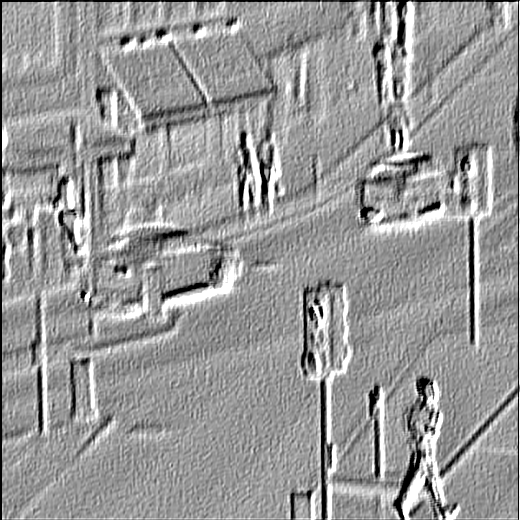}}\,
	\subfigure[]{\includegraphics[width=\tempwidth]{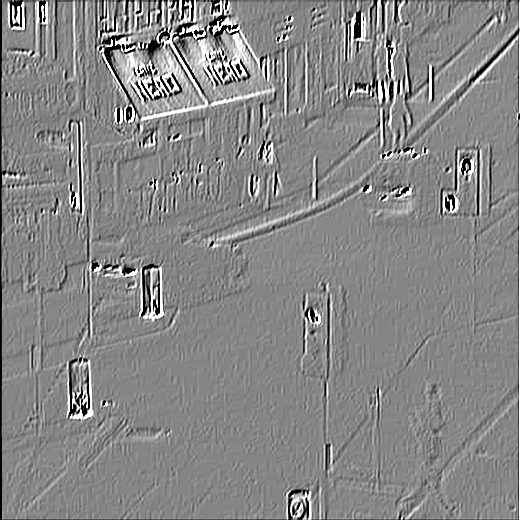}}\,
	\subfigure[]{\includegraphics[width=\tempwidth]{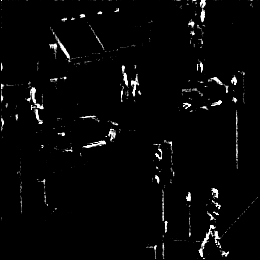}}\,
	\subfigure[]{\includegraphics[width=\tempwidth]{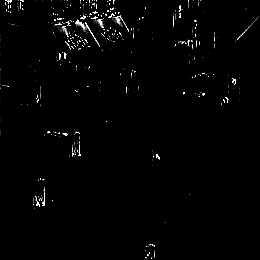}}
	\centering
	\caption{The features of source images through VGG19 network. (a) Infrared image; (b) Visible image; VGG19 learned feature maps from (c) Block1\underline{\hspace{0.5em}}conv1; (d) Block1\underline{\hspace{0.5em}}conv1; (e) Block1\underline{\hspace{0.5em}}conv2; (f) Block1\underline{\hspace{0.5em}}conv2; (g) Block2\underline{\hspace{0.5em}}conv1; (h) Block2\underline{\hspace{0.5em}}conv1.}
	%\textbf{Fig.2.}~~the features of source images through VGG19 network.
	\label{fig:vgg19}
\end{figure}

VGG19 can learn hierarchical features from pixel and edge-related low-level features to semantic related abstract high-level features. However, from the point of view of integrating multi-images, image fusion pays more attention on the low level features learned from bottom layers. To demonstrate this effect, some VGG19 learned feature maps of \dbname{``street''} image are shown in \autoref{fig:vgg19}, where \autoref{fig:vgg19}(c) shows the first layer learned features which are mainly edge-related features from \dbname{``street lamp''} in infrared image and \autoref{fig:vgg19}(d) shows the corresponding features of  \dbname{``billboard''} in visible image. Then the second layer can learn some complicated features, \autoref{fig:vgg19}(e) shows some detail information like the  \dbname{``street lamp''} and the texture of  \dbname{``people''} in the infrared image, and \autoref{fig:vgg19}(f) shows the texture feature of  \dbname{``words on billboard''} in the visible image. The third layer demonstrates more edge combo-features and some semantic features as shown in \autoref{fig:vgg19}(g) and \autoref{fig:vgg19}(h). High-level layers tend to learn higher level semantic features, hence we just transfer the first three convolutions and the following pooling layers of VGG19 into JCAE.

The weights of the first three convolution layers of VGG19 network are used to initialize the encoder layer of the joint convolution auto-encoder network. The pre-trained VGG19 on the ImageNet data set is used to speed up the loss dropping in training stage and improve feature learning abilities. Moreover, VGG19 requires input-shape with three channels, however, infrared and visible images are usually single-channel images, in this study, concatenating along channel dimensions is employed to match the input of VGG network. With these pre-trained weights, JCAE enhances the power of feature learning by fine-tuning on only a small amount of infrared/visible paired images. 

\subsection{JCAE-based fusion rules} \label{sec:fusion_rule}
Once JCAE can reconstruct the input multi-modal source images, it indicates that the feature maps of the hidden layer can be considered as another explanation of the input image pair. So we can design fusion rule in these feature spaces. As discussed in subsection 2.2, the common branch and the private branch have their individual feature learning characteristics, thus we design fusion rules separately according to different branches.

\subsubsection{	Feature fusion of private branches.}
Due to the fact that infrared private features and corresponding visible private features are mutually complementary, for fusing these features with complementary relationship, the choose-max strategy is more appropriate. We formulate this strategy as equation (\ref{eq:12}), where, $F_A^m$ and $F_B^m$  denote the $m^{th}$ feature maps encoded from infrared and visible image respectively and $F_S$ denotes the corresponding fused feature map, $(x, y)$ denotes position in 2D-shape of the feature map. 
\begin{equation}
{F_S}(x, y) = \left\{ {\begin{array}{*{20}{c}}
	{F_A^m(x, y)}&{F_A^m(x, y)> F_B^m(x, y)}\\
	{F_B^m(x, y)}&{F_A^m(x, y)< F_B^m(x, y)}
	\end{array}} \right.
\label{eq:12}
\end{equation}

\subsubsection{Feature fusion of common branches.}
Unlike the private features, common features show the character of the complementary relationship on some feature maps with low activities, on the other hand, they show the character of the redundant relationship on some maps with more activities. To better distinguish feature, we denoted two activity measurements by ${L_K^m}$ and ${C_k}(x,y)$, $M$ represents the total number of feature maps and $m$ stands for one layer feature map. Let $F_k^{1:M}(x,y)$ denote the output of the last layer of the encoder layer, $L_K^m$ is the sum of $\left\{ {F:F_K^m(x,y) \ne 0,K \in \left\{ {A,B} \right\}} \right\}$ which denote the layer-related activity measurement of a feature map. When ${L_K^m}$ is less than a threshold $T$, the choose-max rule is employed, otherwise, the weight-averaging rule is employed, in this study, $T=length*width*3/5$. We then denote ${C_k}(x,y)$ as location-related activity measurement, thus the two weights can be formulated as ${w_1} = {{C_A}(x,y)}/\left({{{C_A}\left(x,y\right)}+{{C_B}\left(x,y\right)}}\right)$ and ${w_2} = {{C_B}(x,y)}/\left({{{C_A}\left(x,y\right)}+{{C_B}\left(x,y\right)}}\right)$, 
where the activity measurement level is calculated by ${C_K}(x,y) = \sum\nolimits_{m = 1}^M {F_K^m} (x,y)$.
Finally, the fusion strategy of common branches is summarized as:
\begin{equation}
{F_S}(x,y) = \left\{ {\begin{array}{*{20}{c}}
	{\left. {\begin{array}{*{20}{c}}
			Max [&{F_A^m(x,y), F_B^m(x,y)}]\\
			\end{array}} \right.}&{{L_K^m}< T}\\
	{{w_1} \cdot {F_A^m}(x,y) + {w_2} \cdot {F_B^m}(x,y)}&{{L_K^m} \geq T}
	\end{array}} \right.
\label{eq:14}
\end{equation}

\section{Experimental results and analysis} \label{sec:exp}
In this section, experiment settings are detailed and experiment results are compared with some existing methods by subjective and objective quality evaluations.

\subsection{Experiment settings}
The evaluation are carried on over the TNO dataset which includes 48 pairs of infrared and visible images and publicly available from  \cite{TNODataset}. All paired images in the feature learning stage were adjusted to gray images with $360*280$ resolution. The pre-trained VGG19 on ImageNet was download from \cite{VGG19weight}. The proposed network was implemented in Keras over TensorFlow, the optimizer was set as ADAM and the learning rate was set as $3\times10^{-4}$, the environment ran on Ubuntu 16.04 equipped with GTX 1080Ti, i7-6850k and 32G RAM.

The proposed method was compared with some typical traditional image fusion methods including method based on guided filtering (GFF) \cite{2013Image}, Laplacian pyramid based image fusion method(LPSR)  \cite{2015ALPSR}, TV-L based image fusion method (GTF) \cite{Ma2016Infrared}, as well as some deep learning based state-of-the-art methods including deep convolution sparse representation based image fusion method(CSR) \cite{Chen2009Research}, convolution neural network based image fusion method(MFCNN) \cite{Liu2017Multi}, and Deepfuse method (Deepfuse) \cite{Prabhakar2017DeepFuse}, Image fusion based on generative adversarial network based method (FusionGAN) \cite{2019FusionGAN}, densenet-based method (Densefuse)  \cite{Hui2018DenseFuse} and the fusion method driven by big data (IFCNN) \cite{2020IFCNN}.  The parameters of these methods are set according to the relevant references.

\subsection{Loss function parameter settings}

\begin{figure}[!htbp]
	\addtolength{\abovecaptionskip}{-5pt}
	\addtolength{\belowcaptionskip}{-10pt}	
	\def\tempwidth{0.20\textwidth}
	\tiny
	\centering
	\subfigure[]{\includegraphics[width=\tempwidth]{source-ir42.png}}\,
	\subfigure[]{\includegraphics[width=\tempwidth]{source-vis42.png}}\,
	\subfigure[]{\includegraphics[width=\tempwidth]{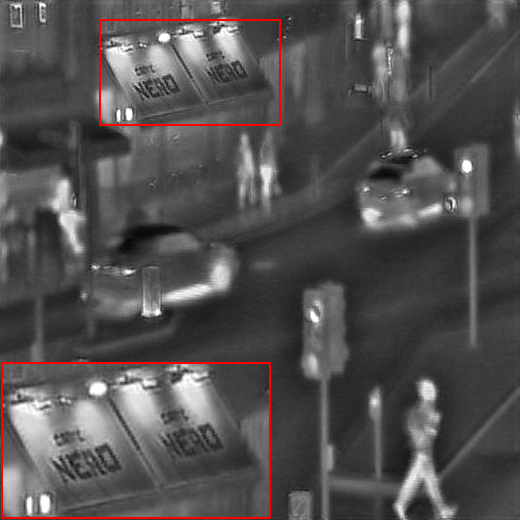}}\,
	\subfigure[]{\includegraphics[width=\tempwidth]{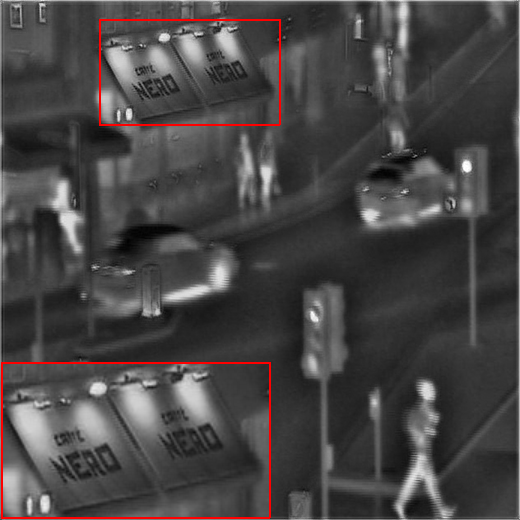}}
	\vspace{-2mm}
	\\
	\subfigure[]{\includegraphics[width=\tempwidth]{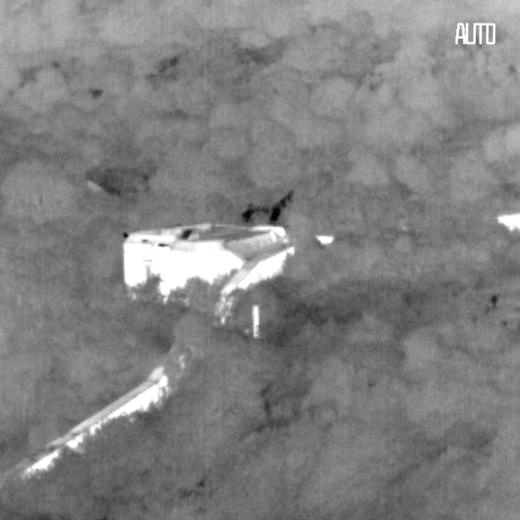} }\,
	\subfigure[]{\includegraphics[width=\tempwidth]{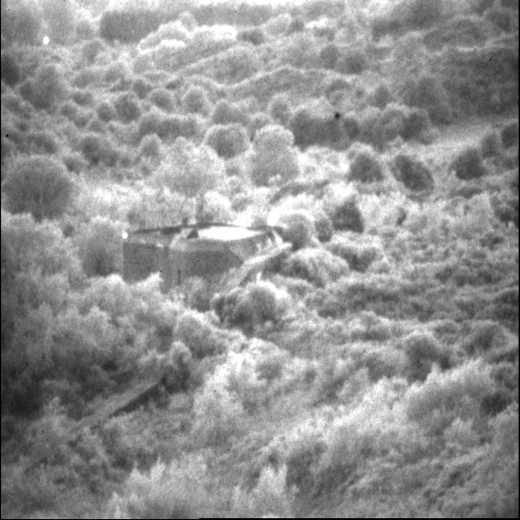}}\,
	\subfigure[]{\includegraphics[width=\tempwidth]{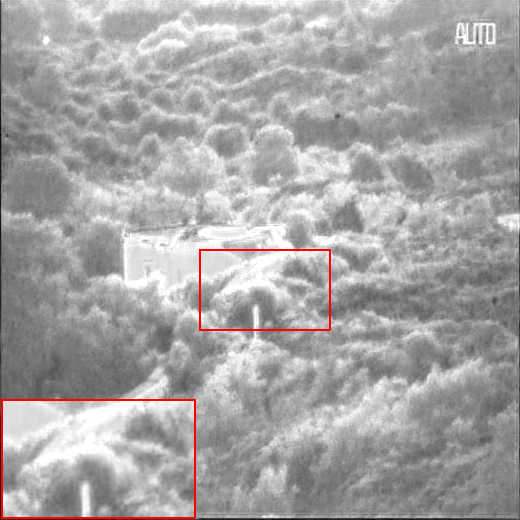}}\,
	\subfigure[]{\includegraphics[width=\tempwidth]{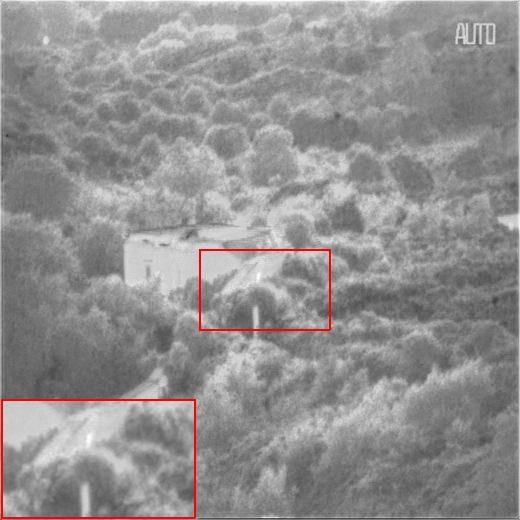}}%
	\caption{Comparison of fusion evaluation metrics before and after being added to the loss function. (a) Infrared image \dbname{street}. (b) Visible image \dbname{street}. (c) Fused \dbname{street} with only the MSE loss function.
			(d) Fused \dbname{street} with the multi-loss function. (e) Infrared image \dbname{bunker}.
			(f) Visible image \dbname{bunker}. (g) Fused \dbname{bunker} with only the MSE loss function. (h) Fused \dbname{bunker} with the multi-loss function.}
	%\centering
	\label{fig:multi-loss}	
\end{figure}

In section \ref{sec:loss_func}	, we have introduced the loss function during network training. In this section, we will mainly discuss the effect of adding the quality evaluation function $L_{ssim}$ to the combined loss function and the setting of the parameter $\lambda$. We take two images named \dbname{street} and \dbname{bunker} to demonstrate the fusion comparison between training with only $L_{mse}$ function and with the multi-loss function, their results are shown in \autoref{fig:multi-loss}, where \autoref{fig:multi-loss} (c) use only $L_{mse}$, although the overall image has better brightness information, the characters on the \dbname{billboard} are blurred. When the multi-loss function are introduced, the results not only retain the overall information of the image, but also integrate the structural details of the image, and visually the characters on the billboard are sharp and clear in \autoref{fig:multi-loss}(d). The similar effect can be seen from the fusion results of \dbname{bunker}, in  \autoref{fig:multi-loss}(g), only MSE loss function is applied, although the brightness information of the forest is good, the edge structure of the bunker and the forest are lost. While in \autoref{fig:multi-loss}(h), when multi-loss is used, the edge of the junction between the bunker and the forest is also very clear. Observed from \autoref{fig:multi-loss}, JCAE demonstrates its advantages when the SSIM loss function was introduced. 

\begin{figure}[!htbp]
	\addtolength{\abovecaptionskip}{-5pt}
	\addtolength{\belowcaptionskip}{-10pt}
	\centering
	\includegraphics[width=\textwidth]{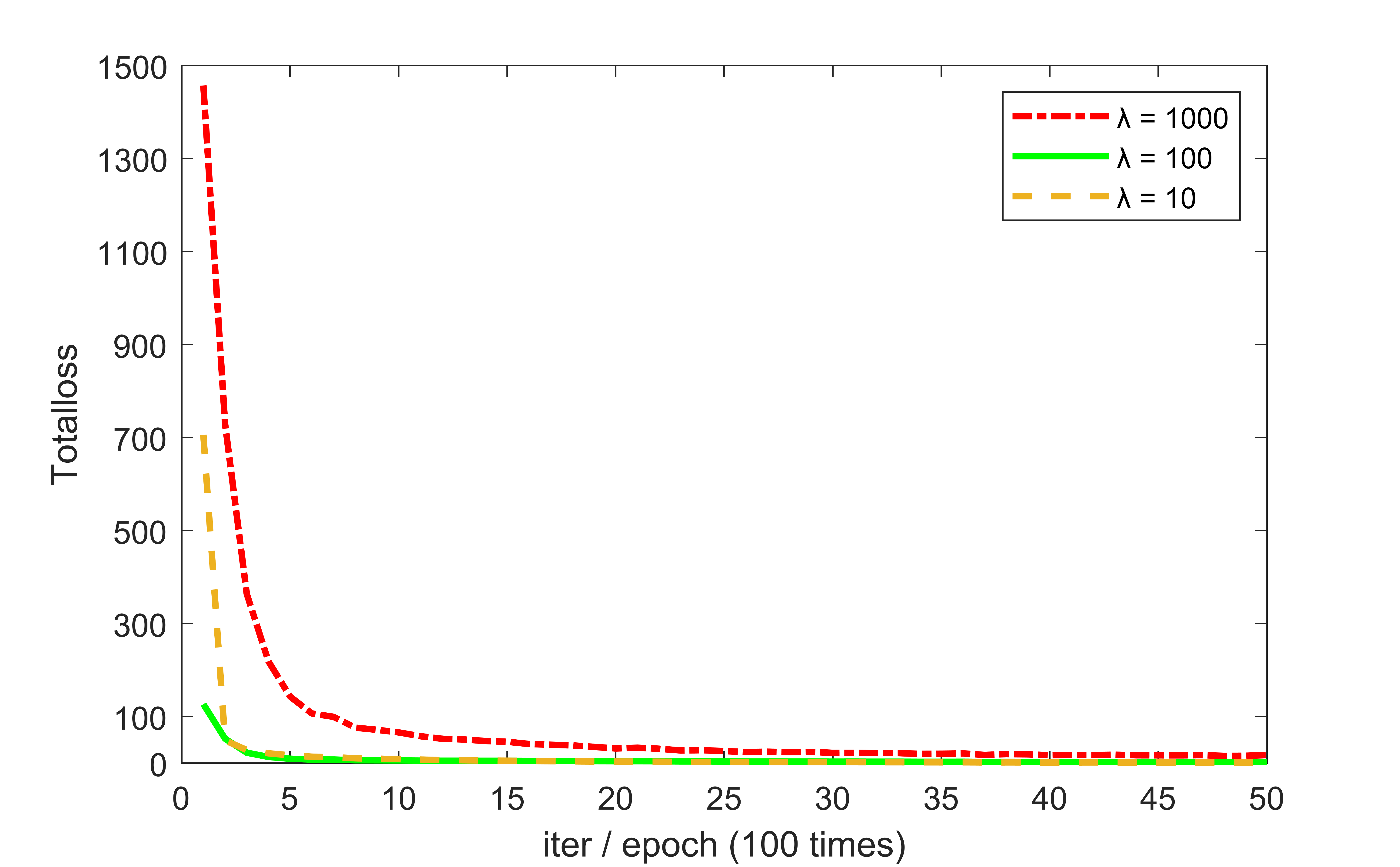}
	\caption{The converging curves of total loss in the training stage with different $\lambda$. $\lambda= 100$ has a faster convergence and declines to a considerable solution. }
	%\textbf{Fig.1.}~~The proposed fusion method
	\label{fig:loss}
\end{figure}

The parameter $\lambda$ balances and scales $L_{mse}$ and $L_{ssim}$. The loss iteration curves with different $\lambda$ are plotted in \autoref{fig:loss}, as can bee seen from these converging curves, once the iterations number exceeds 30 epochs, no matter what value $\lambda$ takes, the loss converge well. However,  the loss curve has a faster convergence when assigning  $100$ to $\lambda$, moreover, this setting exactly adjusts $L_{mse}$ and $L_{ssim}$ to an approximated order of magnitude. So we choose $100$ as the optimal $\lambda$.

\subsection{Private and common features on ``street'' data}
\begin{figure}[htb]
	\addtolength{\abovecaptionskip}{-5pt}
	\addtolength{\belowcaptionskip}{-10pt}	
	\def\tempwidth{0.20\textwidth}
	\tiny
	\centering
	\subfigure[]{\includegraphics[width=\tempwidth]{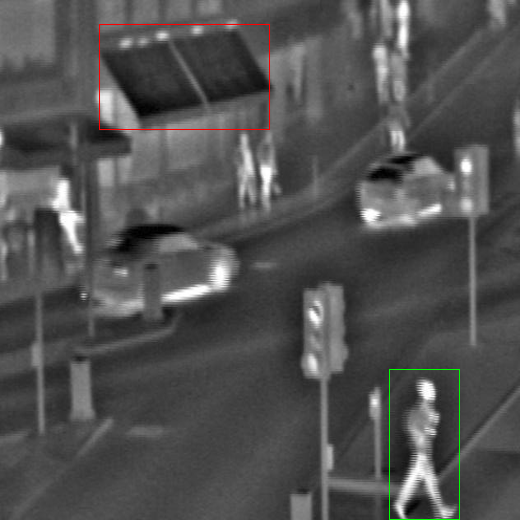}}
	\,
	\subfigure[]{\includegraphics[width=\tempwidth]{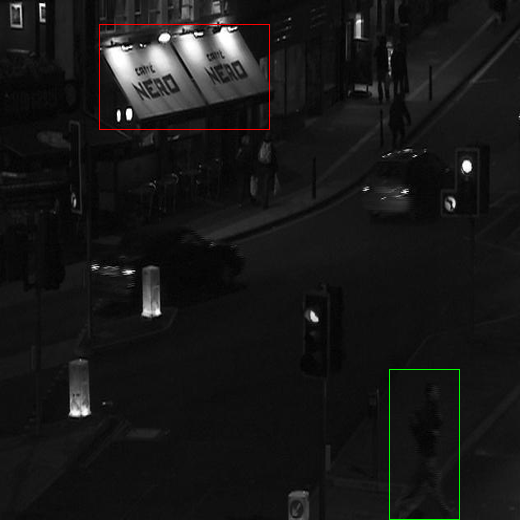}}
	\,
	\subfigure[]{\includegraphics[width=\tempwidth]{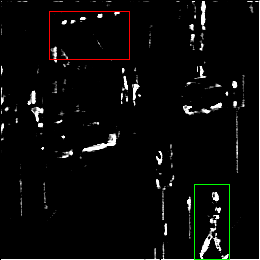}}
	\,
	\subfigure[]{\includegraphics[width=\tempwidth]{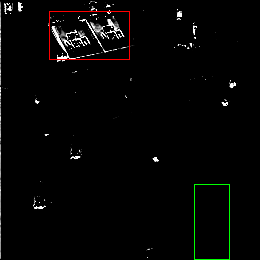}}
	\vspace{-2mm}
	\\
	\subfigure[]{\includegraphics[width=\tempwidth]{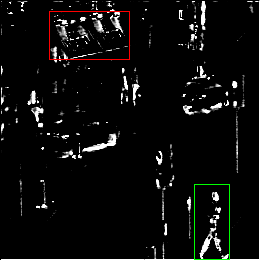} }
	\,
	\subfigure[]{\includegraphics[width=\tempwidth]{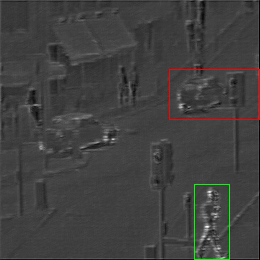}}
	\,
	\subfigure[]{\includegraphics[width=\tempwidth]{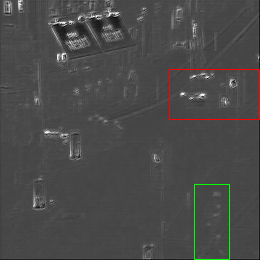}}
	\,
	\subfigure[]{\includegraphics[width=\tempwidth]{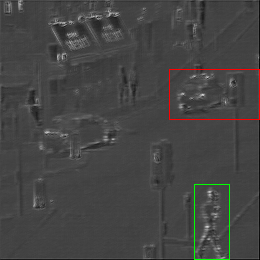}}%
	\caption{Demonstration of the fusion strategy on the feature output by encoder layer. (a) Infrared image. (b) Visible image. (c) One channel feature of the 128 feature maps from the infrared private branch.
		(d) The corresponding visible private branch. (e) The fused feature by  equation \ref{eq:12}.
		(f) One channel of the 128 feature maps from the infrared common branch. (g) The corresponding visible common branch. (h) The fused feature by equation \ref{eq:14}.}
	%\centering
	\label{fig:jcaefeature}	
\end{figure}

We take \dbname{``street''} dataset as an example to demonstrate JCAE’s joint feature learning ability and discuss the fusion rules. The pair of \dbname{ ``street''} images were passed through the trained JCAE and the outputs of encoder layers were visualized. Due to the limitation of space, \autoref{fig:jcaefeature}(c) only shows one channel of the 128 feature maps from the infrared private branch, and \autoref{fig:jcaefeature}(d) shows that of the corresponding visible private branch. \autoref{fig:jcaefeature}(e) shows the fused feature by equation (\ref{eq:12}). \autoref{fig:jcaefeature}(f) \autoref{fig:jcaefeature}(g) and \autoref{fig:jcaefeature}(h) show the corresponding common features and the fused feature by equations (\ref{eq:14}). As can be seen from these figures, private branches can capture specific detail information of source images, the complementary relationship can be seen from the red boxed \dbname{ ``billboard''} and green boxed \dbname{ ``pedestrian''} in \autoref{fig:jcaefeature}(c) and \autoref{fig:jcaefeature}(d), therefore the fused feature map in \autoref{fig:jcaefeature}(e) has more reasonable information. Meanwhile, common branches tend to capture the general shape and structure feature of source images, the \dbname{ ``pedestrian''} activities weak in \autoref{fig:jcaefeature}(f) and does none in \autoref{fig:jcaefeature} (g), moreover, the objects of \dbname{``car''} and \dbname{``road''} both appear in \autoref{fig:jcaefeature} (f) and \autoref{fig:jcaefeature} (g), that means the redundant relationship. As shown in \autoref{fig:jcaefeature}(h), the most useful information is preferably integrated into the fused feature map. These effects are consistent with the purpose of image fusion, extracting and merging the most useful information from multi-images into a single image.

\subsection{Subjective visual evaluation}
\begin{figure}[htb]
	\addtolength{\abovecaptionskip}{-5pt}
	\addtolength{\belowcaptionskip}{-10pt}
	\def\tempwidth{0.16\textwidth}
	\tiny
	\centering
	\subfigure[]{\includegraphics[width=\tempwidth]{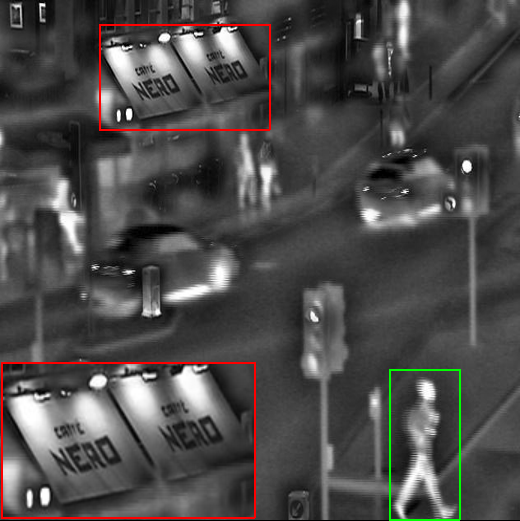}}
	\,
	\subfigure[]{\includegraphics[width=\tempwidth]{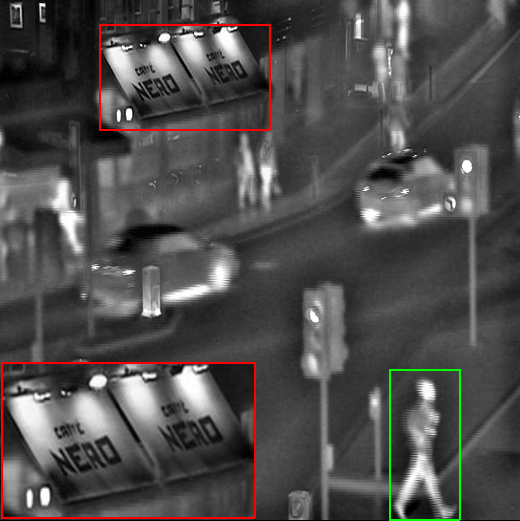}}
	\,
	\subfigure[]{\includegraphics[width=\tempwidth]{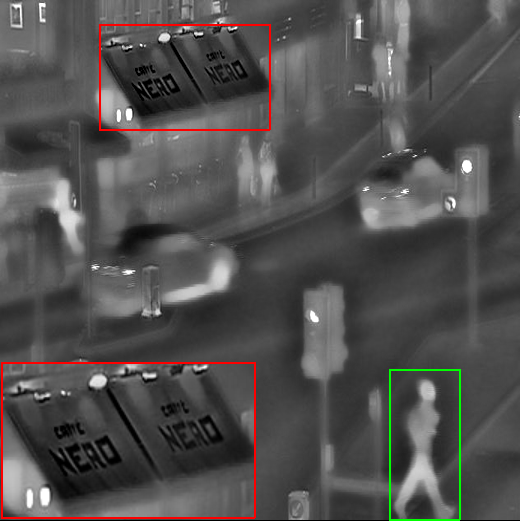}}
	\,
	\subfigure[]{\includegraphics[width=\tempwidth]{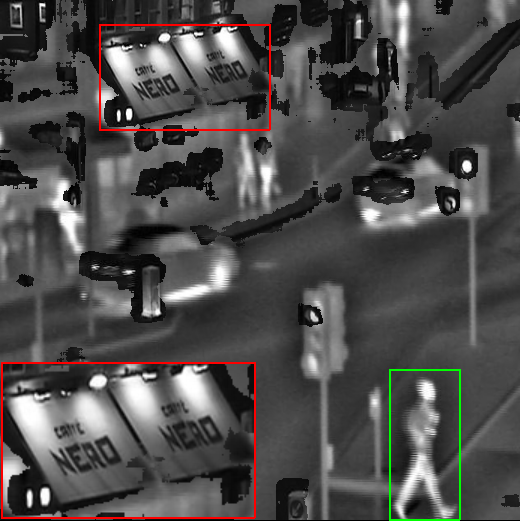}}
	\,
	\subfigure[]{\includegraphics[width=\tempwidth]{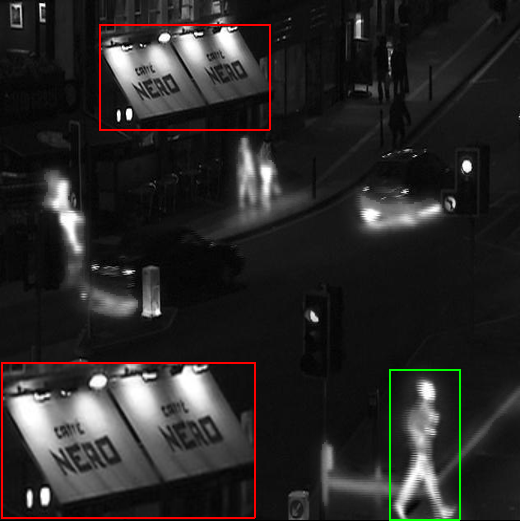}}
	\\
	\vspace{-2mm}
	\subfigure[]{\includegraphics[width=\tempwidth]{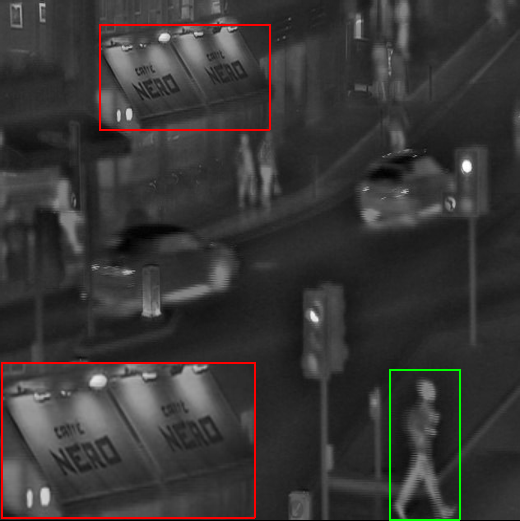}}	
	\,
	\subfigure[]{\includegraphics[width=\tempwidth]{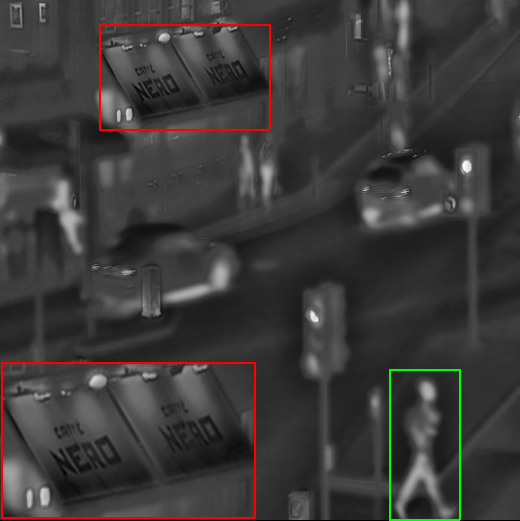}}
	\,
	\subfigure[]{\includegraphics[width=\tempwidth]{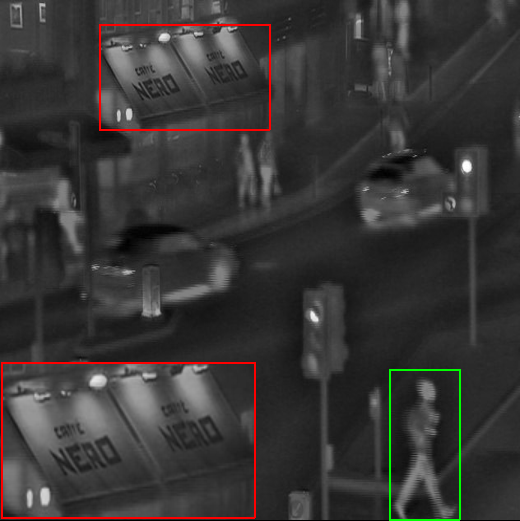}}
	\,
	\subfigure[]{\includegraphics[width=\tempwidth]{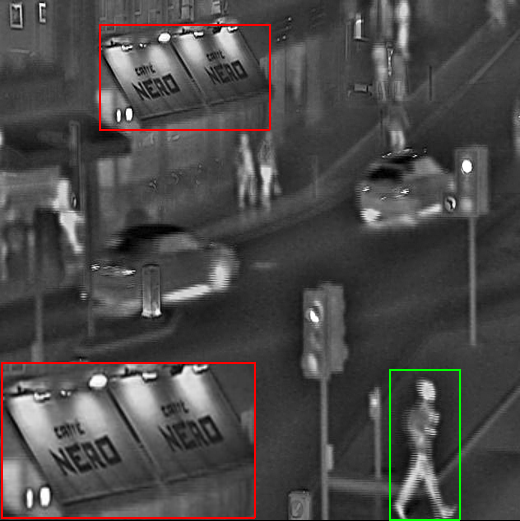}}
	\,
	\subfigure[]{\includegraphics[width=\tempwidth]{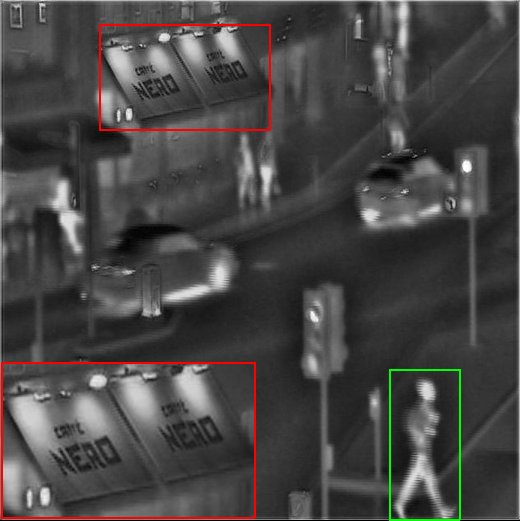}}
	\centering
	\caption{The experiment results of \dbname{``street''} image. Results of (a) GFF. (b) LPSR. (c) GTF. (d) CSR. (e)MFCNN. (f) Deepfuse. (g) GAN. (h) Densefuse. (i) IFCNN. (j) Proposed method.}
	\label{fig:zhuguanxiaoguo}
\end{figure}

To subjectively compare the fusion results between our proposed method and other state-of-art methods, their fusion results are shown in \autoref{fig:zhuguanxiaoguo}. To better observe the fusion results, we outline the billboard with red boxes and pedestrians with green boxes. There are some black block noises in the results of GFF and LPSR methods. The results of GTF have low contrast and brightness, and the information on the billboard is unclear. In the result of MFCNN, due to the existence of the decision diagram, the points with larger pixels in the infrared image are imported, and almost all the information in the visible light is lost. The CSR method generates a lot of noise and the fusion image is contaminated. Deepfuse, GAN, and Densefuse extract the information in a pair of pictures well, but the lack of fusion rules makes these three methods unable to perform a good fusion of features, and the image brightness of the three is dark. IFCNN is a big data-driven fusion method, which benefits from a shallower network layer, and the image information is better preserved. Our method has achieved better visual effects, better contrast, better brightness and detailed information due to separated feature representation.

We also show the comparison fusion results on other pairs of images (\dbname{Fog, Forest, Building, House, Boat, Solider and  Umbrella}) in \autoref{fig:undefine}. Our proposed method also demonstrates good fusion results. The ability of feature encoding and decoding of JCAE contributes to the image reconstruction effect, moreover, the ability of feature learning with redundant and complementary and these features based fusion rules contribute to the fusion effect.
\begin{table}[htb]
	%\footnotesize
	\scriptsize
	\setlength\tabcolsep{6pt} 
	\centering
	%\textbf{Table 2}~~Objective evaluation index comparison of \emph{ ``street''} image adopting different fusion methods.The best values for quality metrics are indicated in bold and red while the second-best values are indicated in italic.\\
	\caption{Objective evaluation comparison on TNO dataset adopting different fusion methods. The bold values indicate the best winner. The value is expressed as $Average \pm Significance$.}
\begin{tabular}{c c c c c c }
	\hline \\[-6pt]
	\textbf{Methods} & \textbf{MI} & \textbf{CC}& \textbf{Q$_{cv}$}& \textbf{SCD}& \textbf{SSIM} \\[-6pt]\tabularnewline
	\hline \\[-6pt]
	%{\color{red}
	%\textbf{
	GFF & 4.2452$\pm 0.00$ &0.6288$\pm 0.00$ & 786.8079$\pm 0.3$ & 1.1939$\pm 0.00$ & 0.6793$\pm 0.01$  \\[-6pt]\tabularnewline
	\hline \\[-6pt]
	LPSR &2.4369$\pm 0.05$ &0.6445$\pm 0.02$&689.0903$\pm 0.94$&1.4009$\pm 0.07$&0.6971$\pm 0.15$ 
	\\[-6pt]\tabularnewline
	\hline \\[-6pt]
	GTF &2.7619$\pm 0.87$&0.6388$\pm 0.01$&\textbf{1089.6090}$\pm 0.02$&0.9638$\pm 0.00$&0.6816$\pm 0.02$ \\[-6pt]\tabularnewline
	\hline \\[-6pt]
	CSR &4.9210$\pm 0.00$ & 0.6116$\pm 0.00$& 1000.9480$\pm 0.03$ & 1.1022$\pm 0.00$ &	0.6407$\pm 0.00$ \\[-6pt]\tabularnewline
	\hline \\[-6pt]
	MFCNN &\textbf{5.8857}$\pm 0.00$&0.6647$\pm 0.17$&428.5851$\pm 0.01$& 1.2357$\pm 0.41$& 0.6966$\pm 0.12$ \\[-6pt]\tabularnewline
	\hline \\[-6pt]
	Deepfuse &2.3052$\pm 0.00$ &0.7074$\pm 0.32$ &535.3907$\pm 0.08$ & \textbf{1.8088}$\pm 0.00$ &0.7152$\pm $0.71 \\[-6pt]\tabularnewline
	\hline \\[-6pt]
	GAN &2.4173$\pm 0.02$ &\textbf{0.7245}$\pm 0.04$&952.5548$\pm 0.02$ &1.3945$\pm 0.09$ &0.6409$\pm 0.00$
	\\[-6pt]\tabularnewline
	\hline \\[-6pt]
	Densefuse &2.3035$\pm 0.00$ &0.7077$\pm 0.31$ &528.1753$\pm 0.07$ &1.8027$\pm 0.00$ &0.7200$\pm 0.97$
	\\[-6pt]\tabularnewline
	\hline \\[-6pt]
	IFCNN &2.4928$\pm 0.06$ &0.6912$\pm 0.89$ &372.7647$\pm 0.00$ &1.6822$\pm 0.00$ &0.7164$\pm 0.80$
	\\[-6pt]\tabularnewline
	\hline \\[-6pt]
	Proposed &2.7308 &0.6887 &678.1914 &1.5237 &\textbf{0.7205}\\[-6pt]
	\tabularnewline   
	\hline
	\end{tabular}
	\label{tab:index}
\end{table}

\subsection{Quality evaluation}
%\paragraph
To further compare the fusion quality of the proposed method with that of other methods, several broadly used image fusion quality evaluation metrics were used to evaluate the fusion results, including Mutual information (MI), Correlation coefficient (CC) \cite{MANJUSHA2010Image}, Information evaluation factor(Q$_{cv}$) \cite{Liu2011Objective}, Sum of the correlations of differences (SCD) \cite{V2015A} and Structural similarity metric measure (SSIM) \cite{ssim}. MI indicates the amount of information contained in the source image in the fusion image. CC and SCD indicate the similarity between the fused image and the source image. SSIM measures image loss and distortion according to the fact that the human visual system is sensitive to structural errors and distortion. Q$_{cv}$ measures the amount of edge information transmitted from the original image to the fused image.

We averaged the results of 48 pairs of pictures in the TNO data set, but the average result may cause fraud depending on the quality of a certain picture. To describe results more accurately, we have introduced a significance index. The results of our proposed method and comparative experiments on the TNO dataset can be regarded as a paired-sample t-test, and the correlation between two samples is judged by calculating the difference between each pair of values. When the significance level index is greater than 0.05, it means that there is no difference between the results of our method and the comparison experiment. When the significance level index is less than 0.05, it means that there is a significant difference between the results of our method and the comparison experiment.

\begin{figure*}[!htbp]
	\centering
	\rotatebox{90}{Infrared}
	{\includegraphics[height=0.12\linewidth]{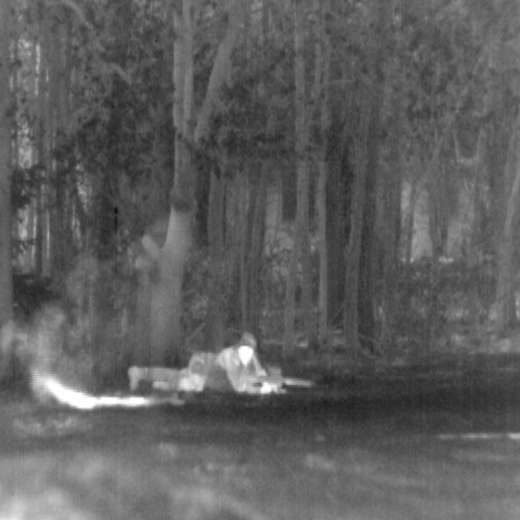}}
	{\includegraphics[height=0.12\linewidth]{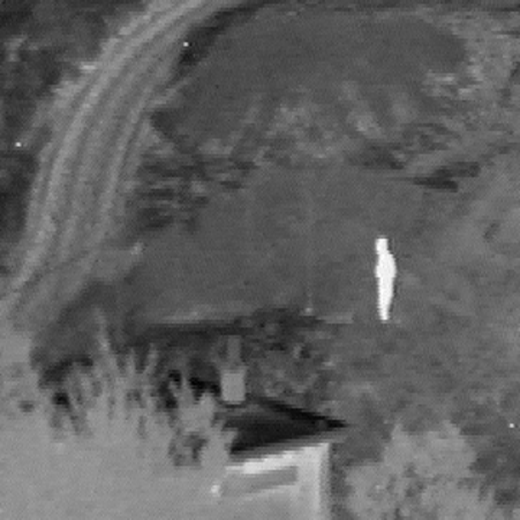}}
	{\includegraphics[height=0.12\linewidth]{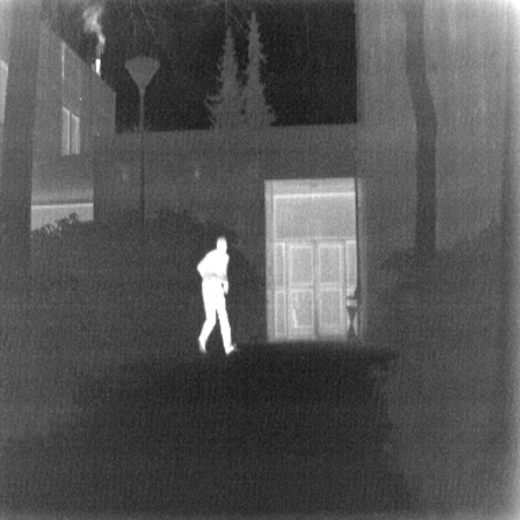}}
	{\includegraphics[height=0.12\linewidth]{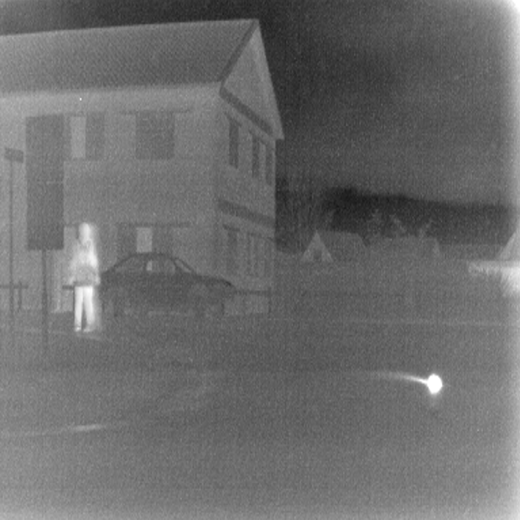}}
	{\includegraphics[height=0.12\linewidth]{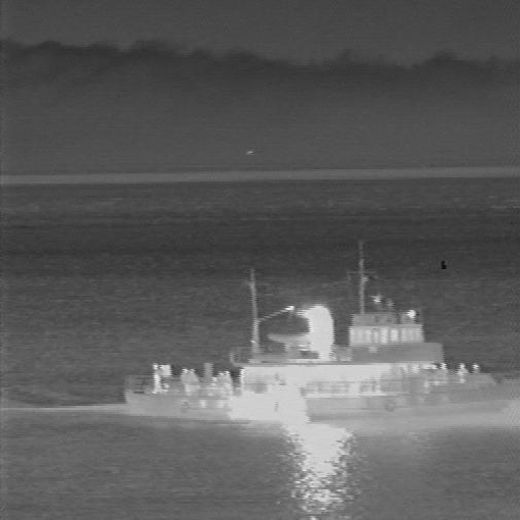}}
	{\includegraphics[height=0.12\linewidth]{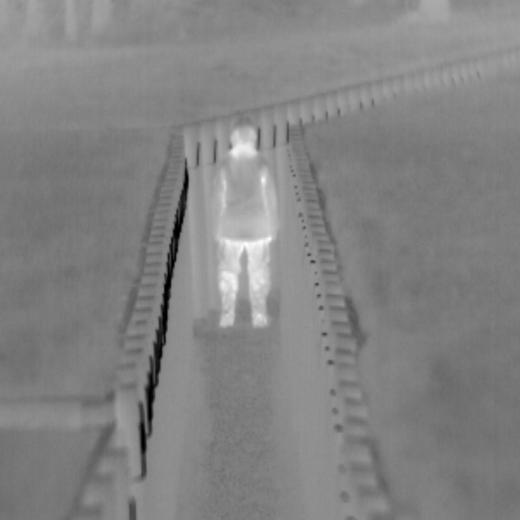}}
	{\includegraphics[height=0.12\linewidth]{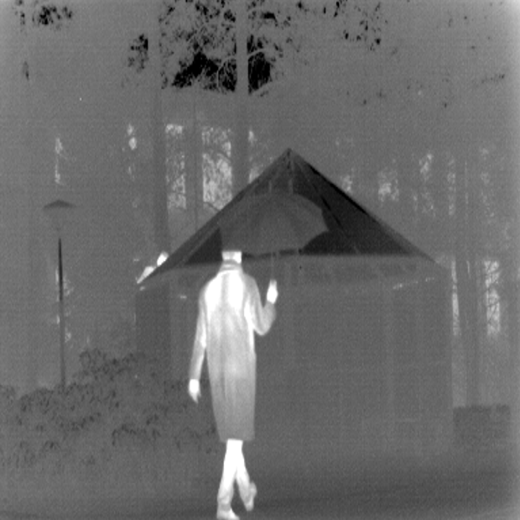}}
	\\
	\rotatebox{90}{Visible}
	{\includegraphics[height=0.12\linewidth]{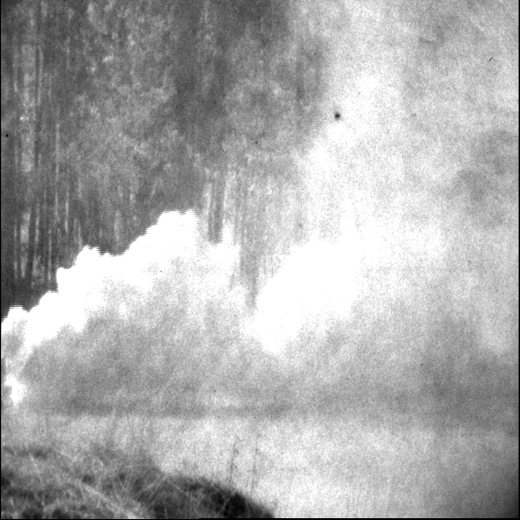}}
	{\includegraphics[height=0.12\linewidth]{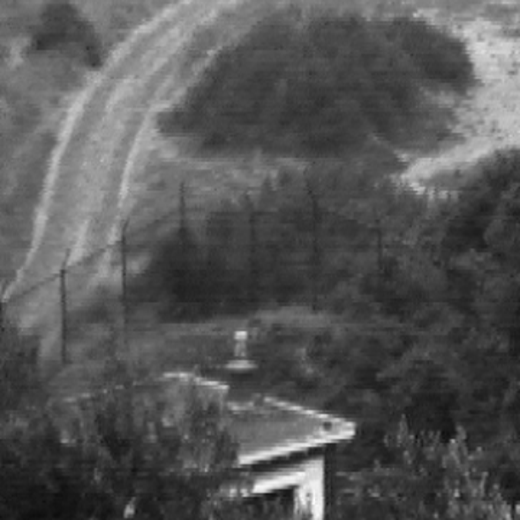}}
	{\includegraphics[height=0.12\linewidth]{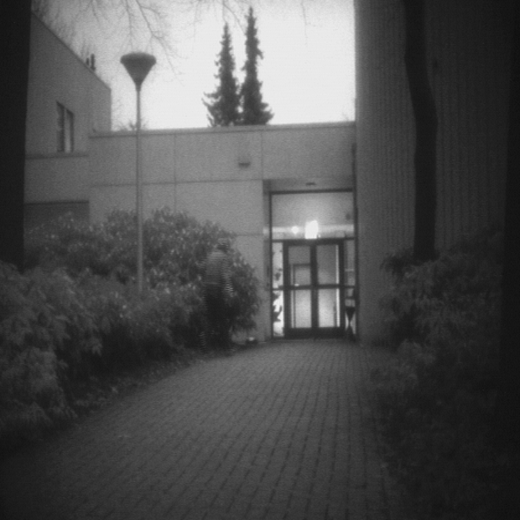}}
	{\includegraphics[height=0.12\linewidth]{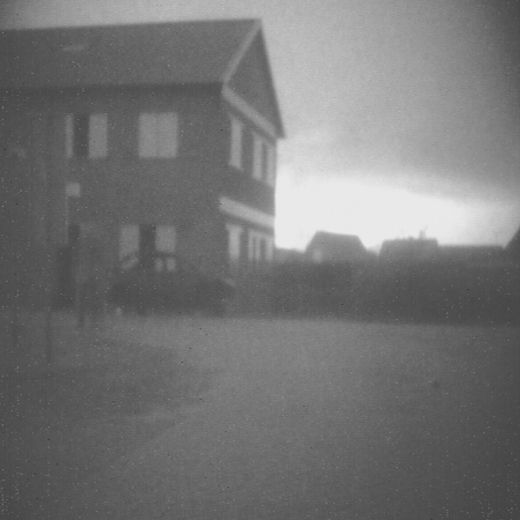}}
	{\includegraphics[height=0.12\linewidth]{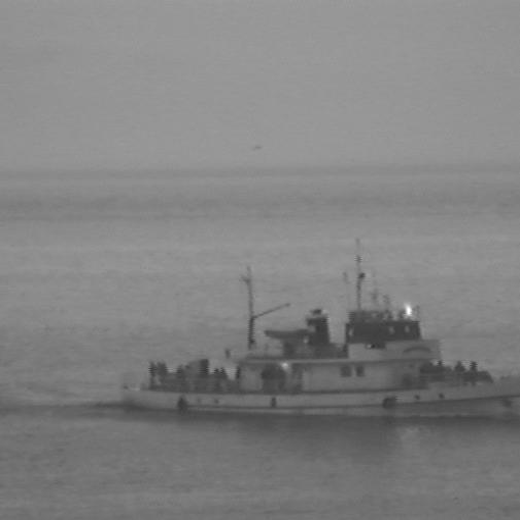}}
	{\includegraphics[height=0.12\linewidth]{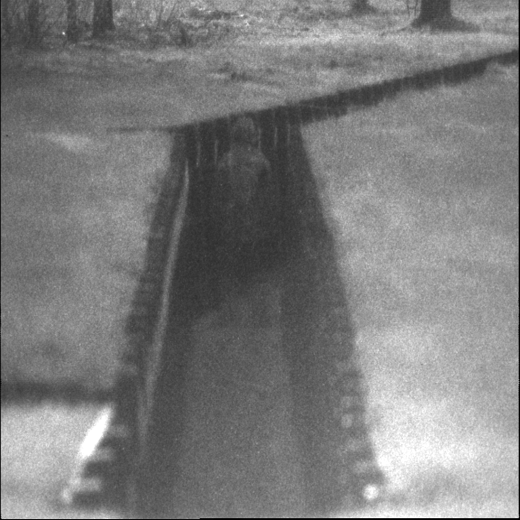}}
	{\includegraphics[height=0.12\linewidth]{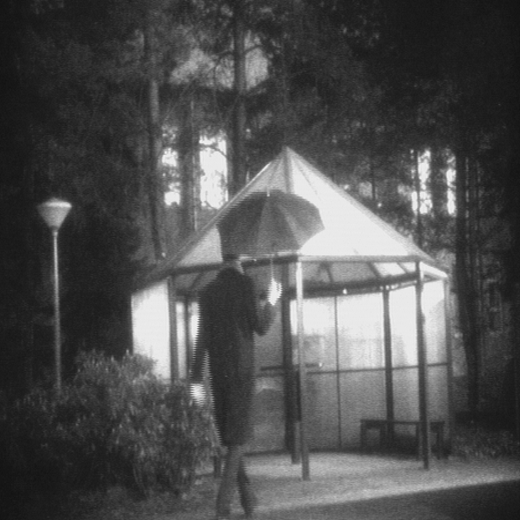}}
	\\
	\rotatebox{90}{GFF}
	{\includegraphics[height=0.12\linewidth]{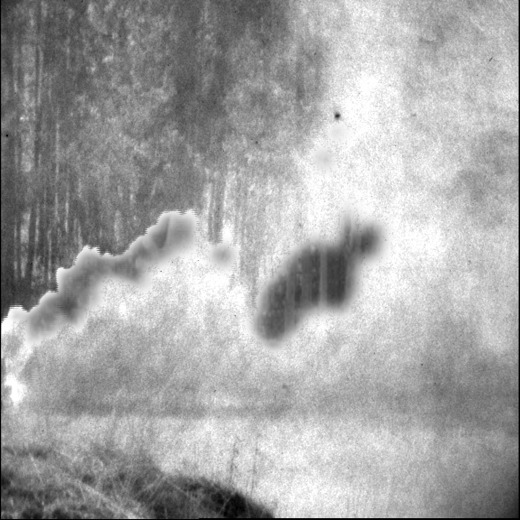}}
	{\includegraphics[height=0.12\linewidth]{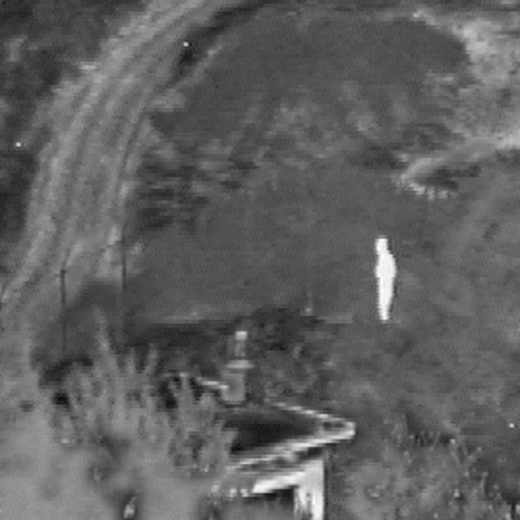}}
	{\includegraphics[height=0.12\linewidth]{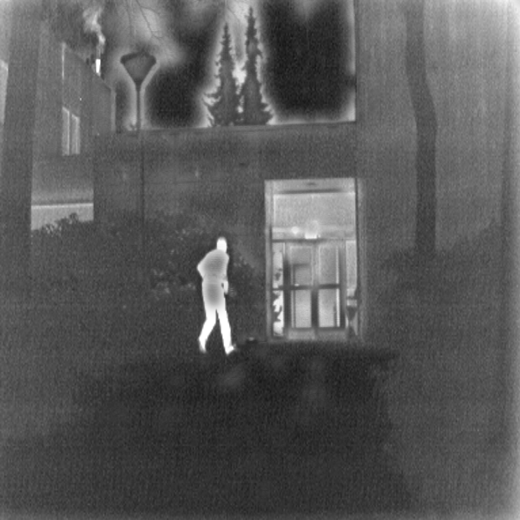}}
	{\includegraphics[height=0.12\linewidth]{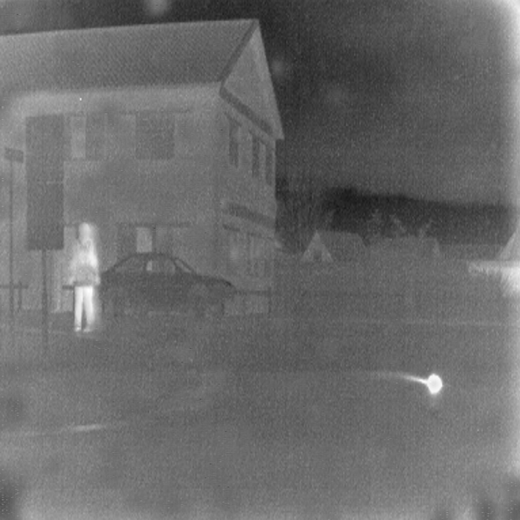}}
	{\includegraphics[height=0.12\linewidth]{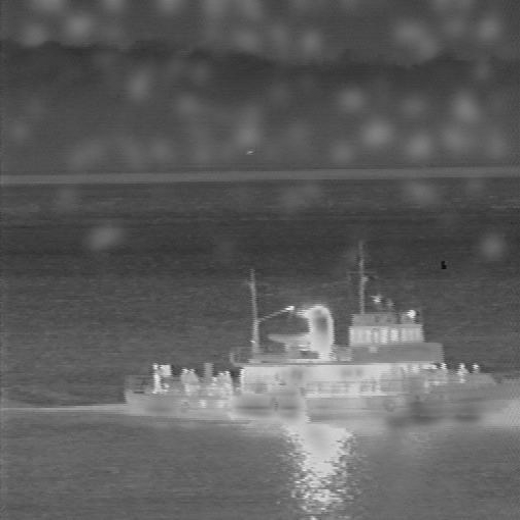}}
	{\includegraphics[height=0.12\linewidth]{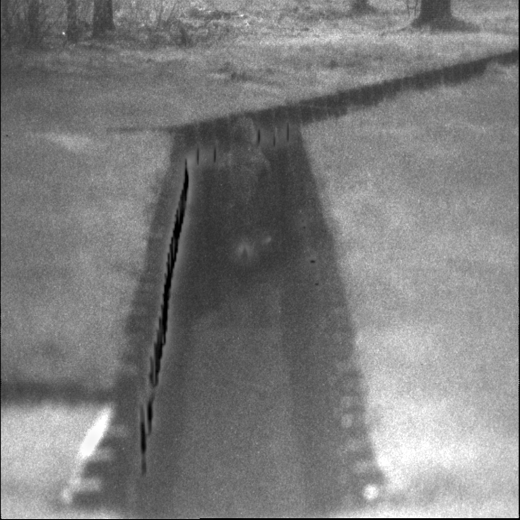}}
	{\includegraphics[height=0.12\linewidth]{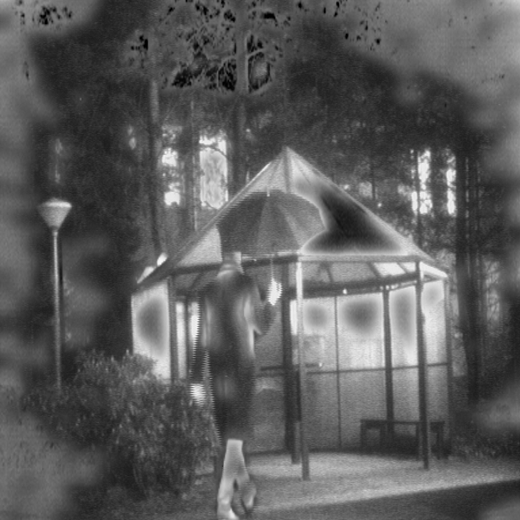}}
	\\
	\rotatebox{90}{LPSR}
	{\includegraphics[height=0.12\linewidth]{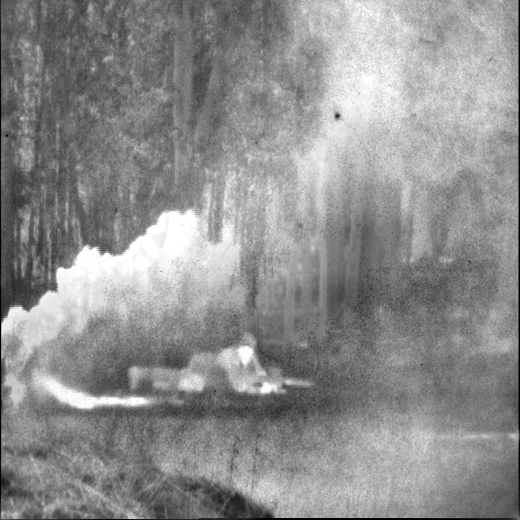}}
	{\includegraphics[height=0.12\linewidth]{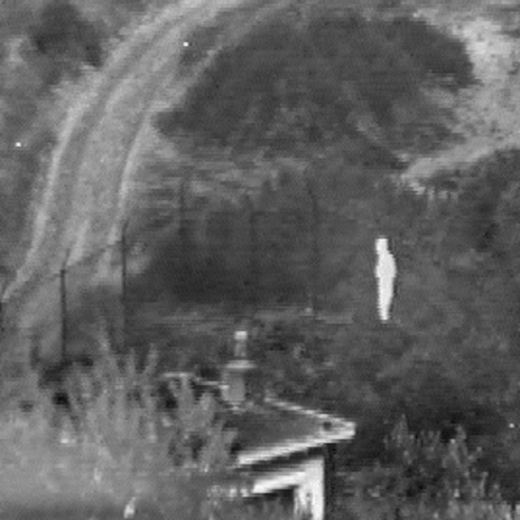}}
	{\includegraphics[height=0.12\linewidth]{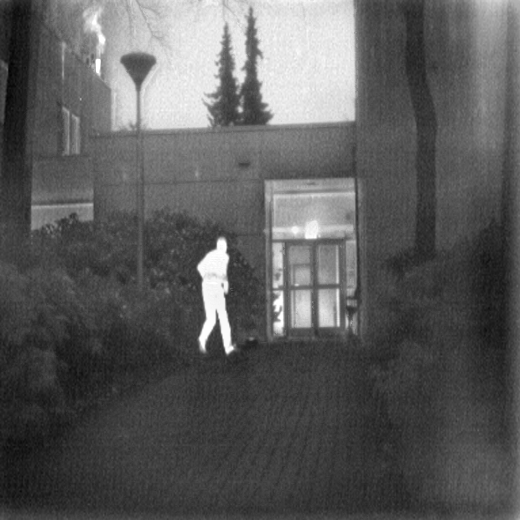}}
	{\includegraphics[height=0.12\linewidth]{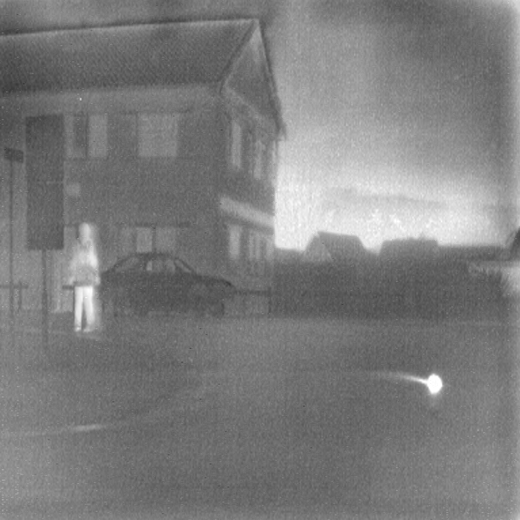}}
	{\includegraphics[height=0.12\linewidth]{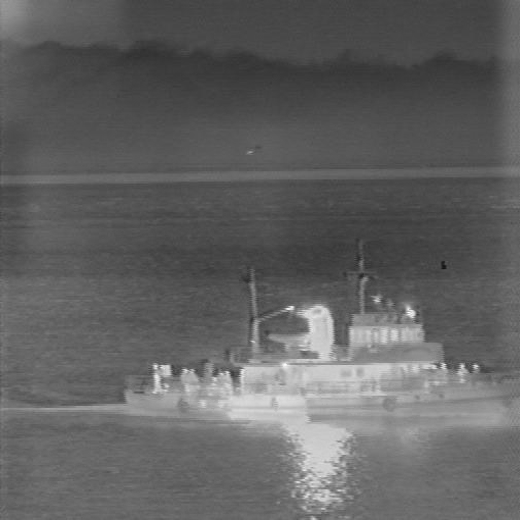}}
	{\includegraphics[height=0.12\linewidth]{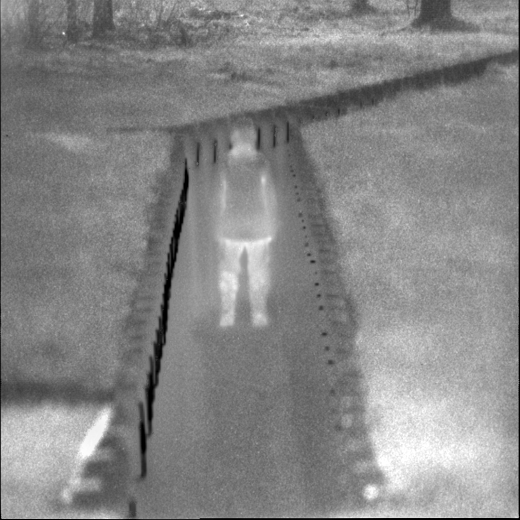}}
	{\includegraphics[height=0.12\linewidth]{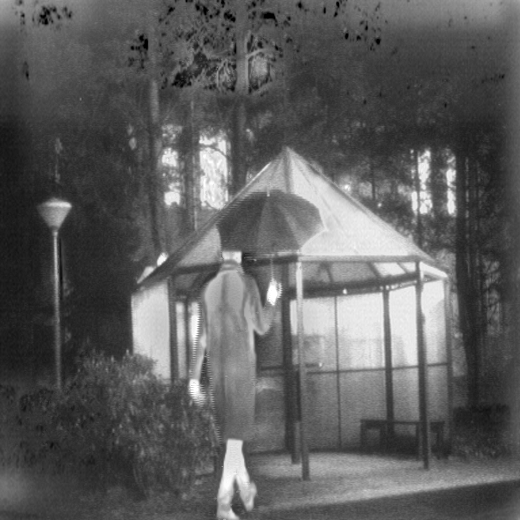}}
	\\
	\rotatebox{90}{GTF}
	{\includegraphics[height=0.12\linewidth]{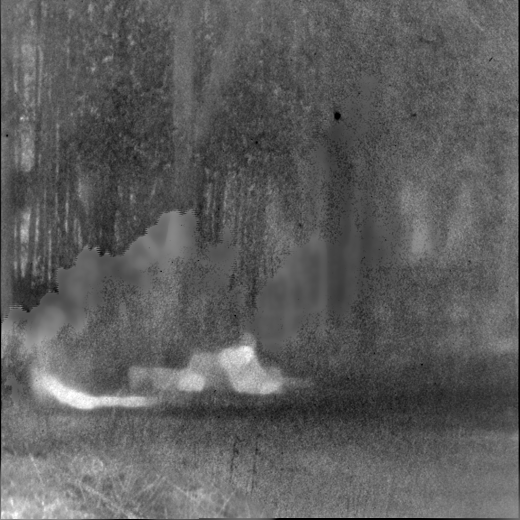}}
	{\includegraphics[height=0.12\linewidth]{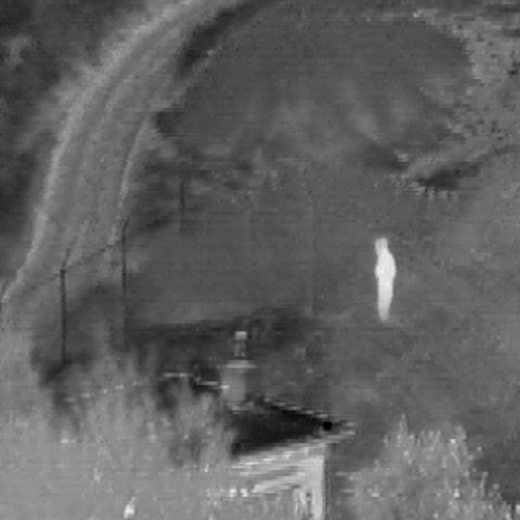}}
	{\includegraphics[height=0.12\linewidth]{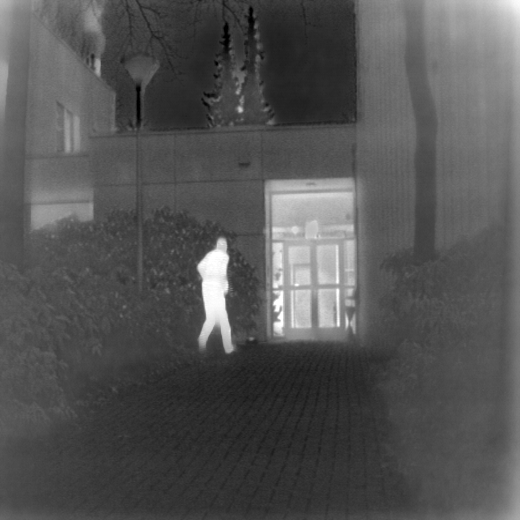}}
	{\includegraphics[height=0.12\linewidth]{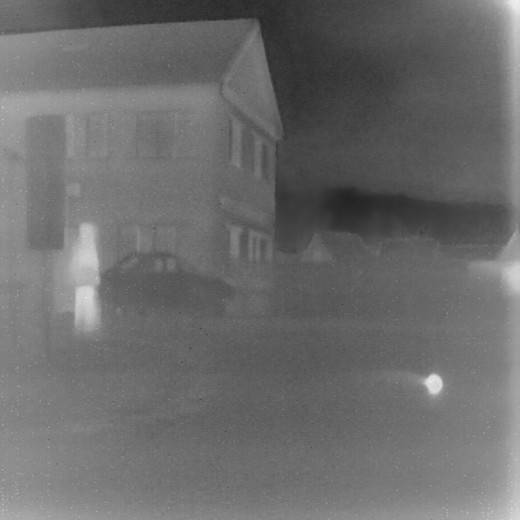}}
	{\includegraphics[height=0.12\linewidth]{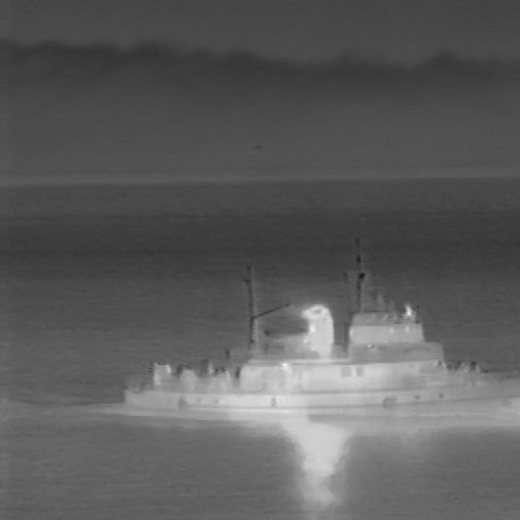}}
	{\includegraphics[height=0.12\linewidth]{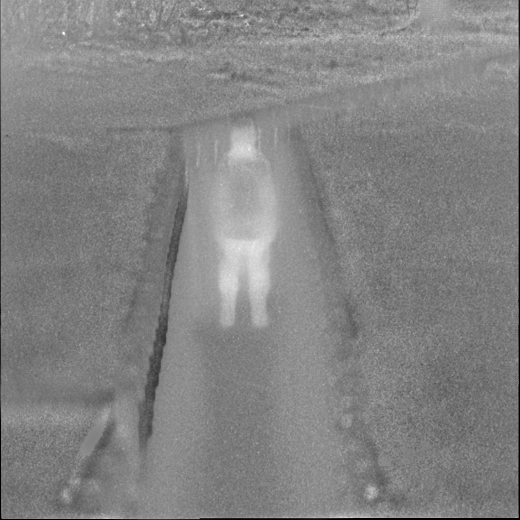}}
	{\includegraphics[height=0.12\linewidth]{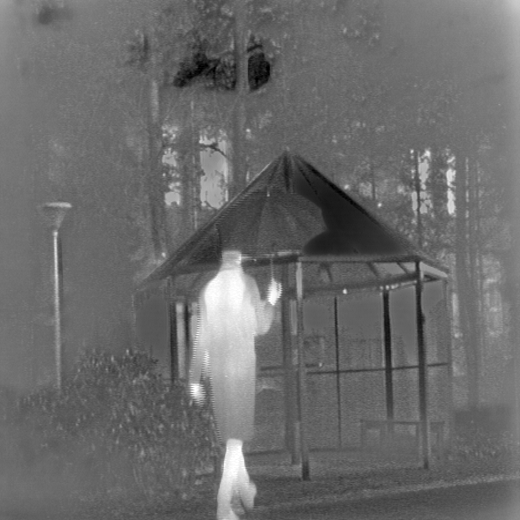}}
	\\
	\rotatebox{90}{CSR}
	{\includegraphics[height=0.12\linewidth]{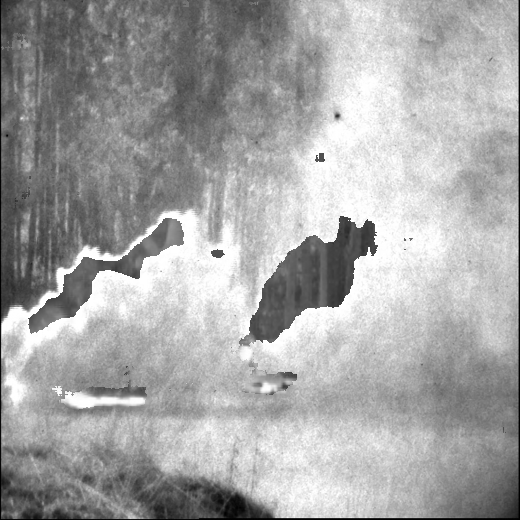}}
	{\includegraphics[height=0.12\linewidth]{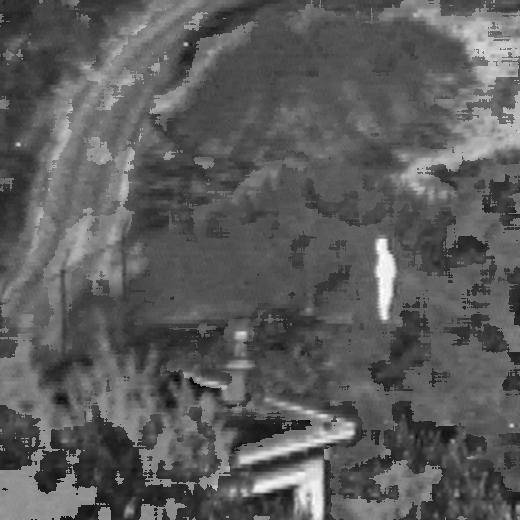}}
	{\includegraphics[height=0.12\linewidth]{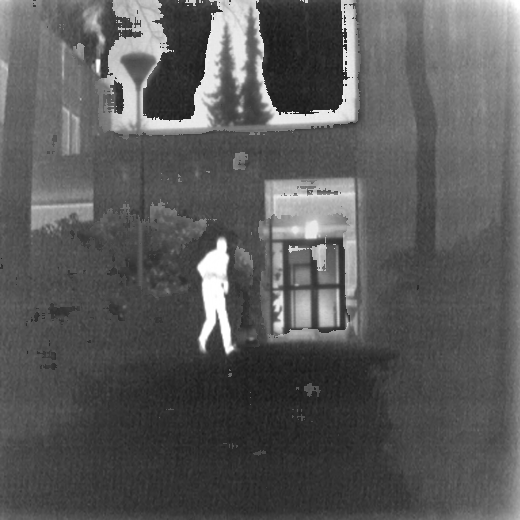}}
	{\includegraphics[height=0.12\linewidth]{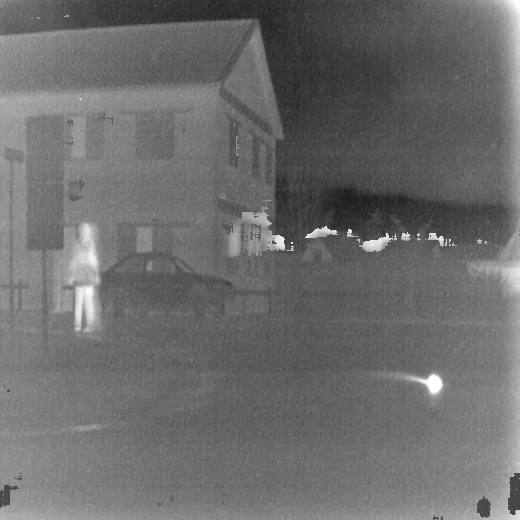}}
	{\includegraphics[height=0.12\linewidth]{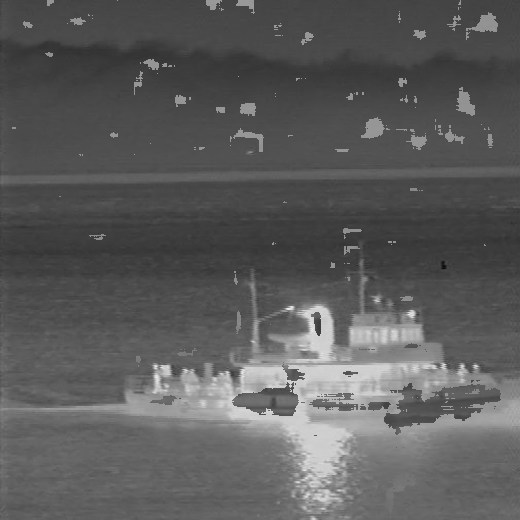}}
	{\includegraphics[height=0.12\linewidth]{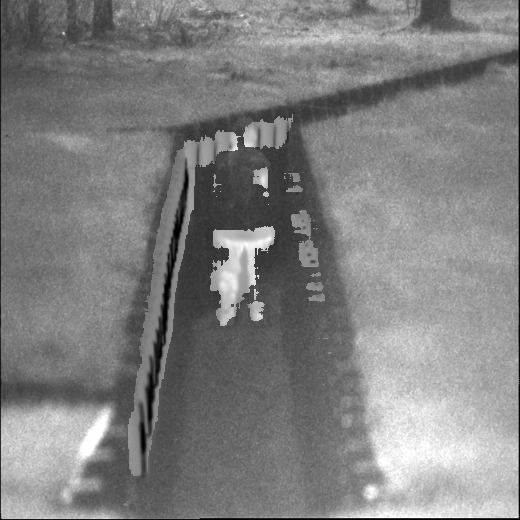}}
	{\includegraphics[height=0.12\linewidth]{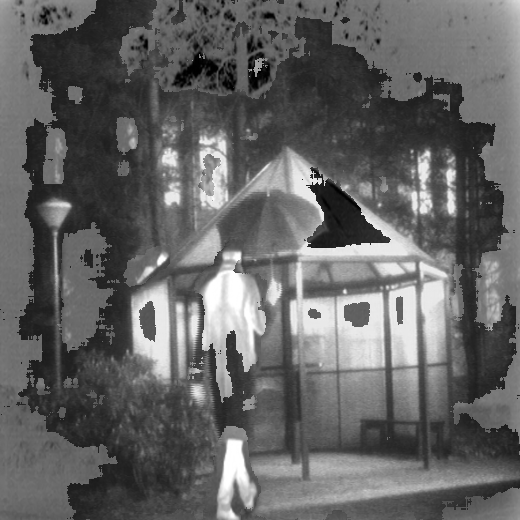}}
	\\
	\rotatebox{90}{MFCNN}
	{\includegraphics[height=0.12\linewidth]{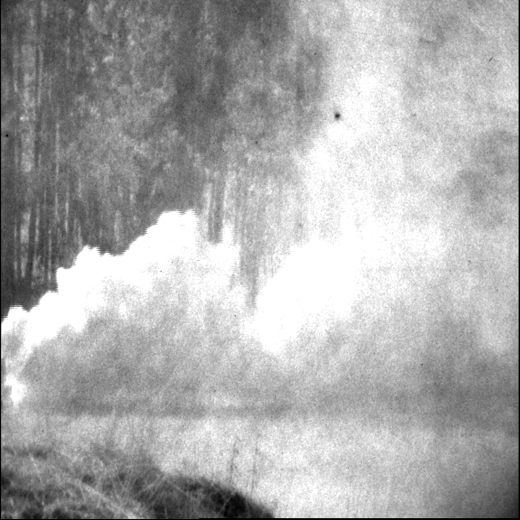}}
	{\includegraphics[height=0.12\linewidth]{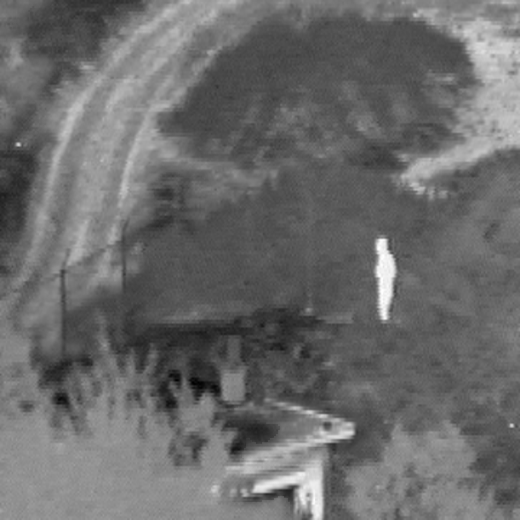}}
	{\includegraphics[height=0.12\linewidth]{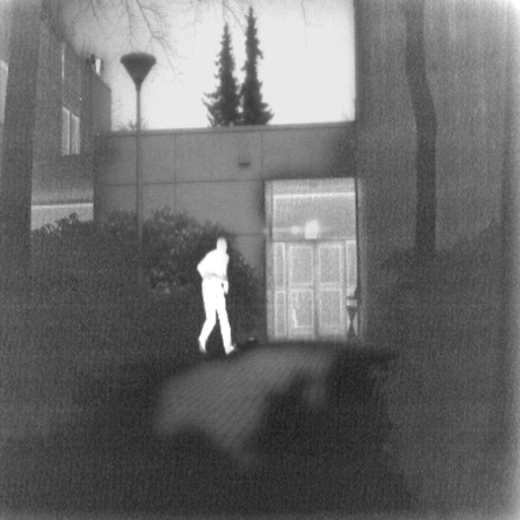}}
	{\includegraphics[height=0.12\linewidth]{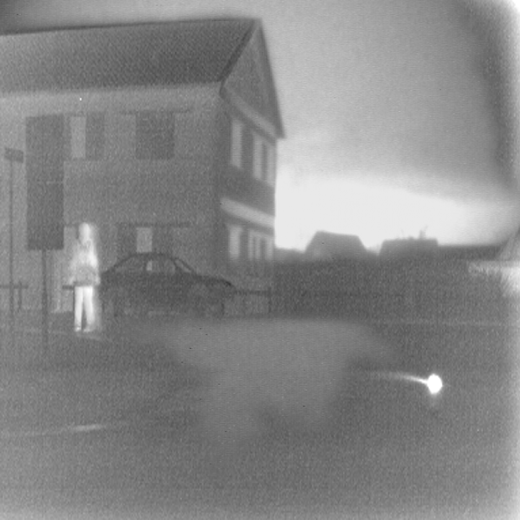}}
	{\includegraphics[height=0.12\linewidth]{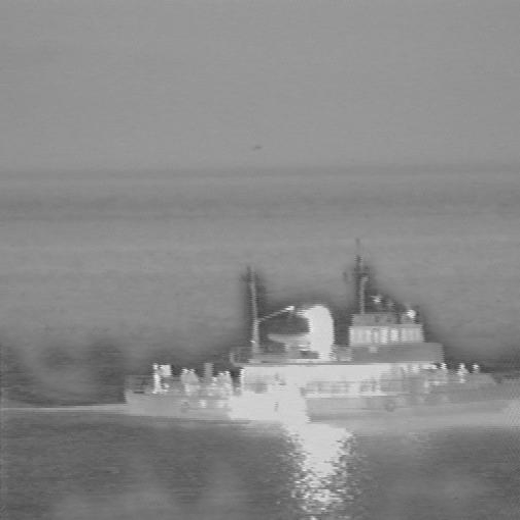}}
	{\includegraphics[height=0.12\linewidth]{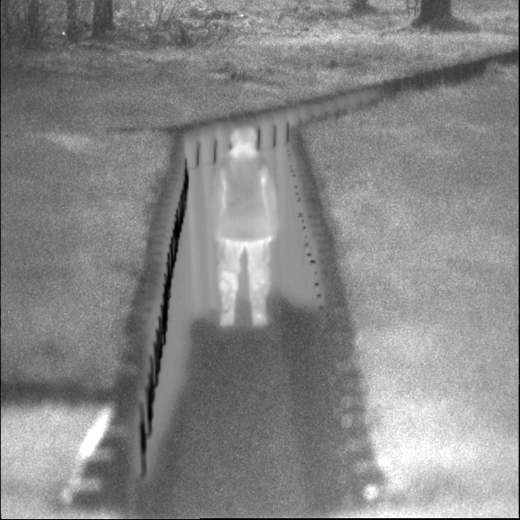}}
	{\includegraphics[height=0.12\linewidth]{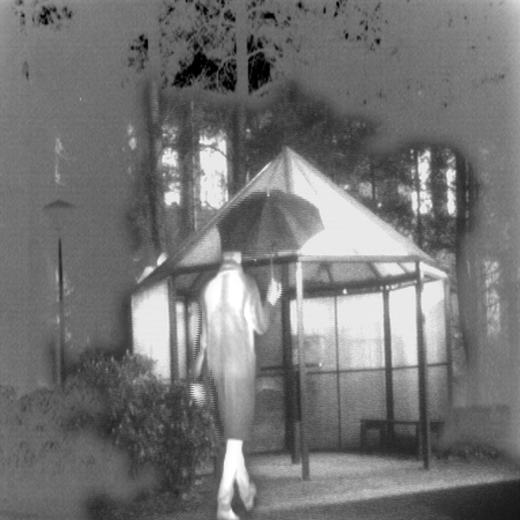}}
	\\
	\rotatebox{90}{Deepfuse}
	{\includegraphics[height=0.12\linewidth]{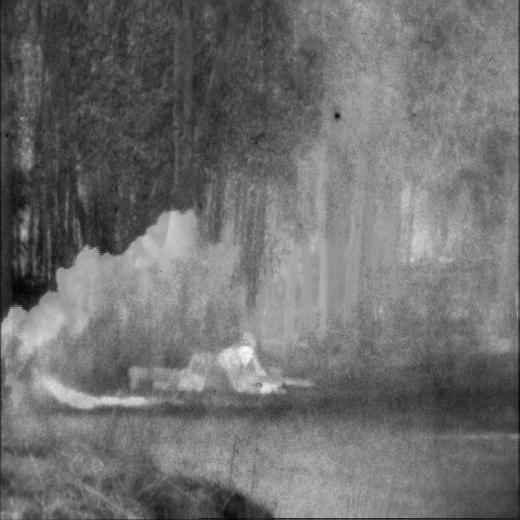}}
	{\includegraphics[height=0.12\linewidth]{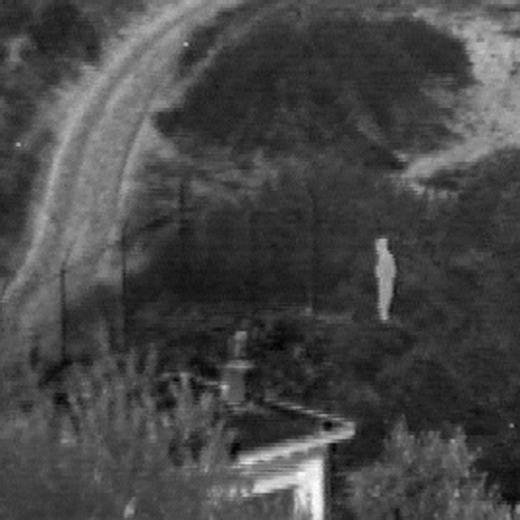}}
	{\includegraphics[height=0.12\linewidth]{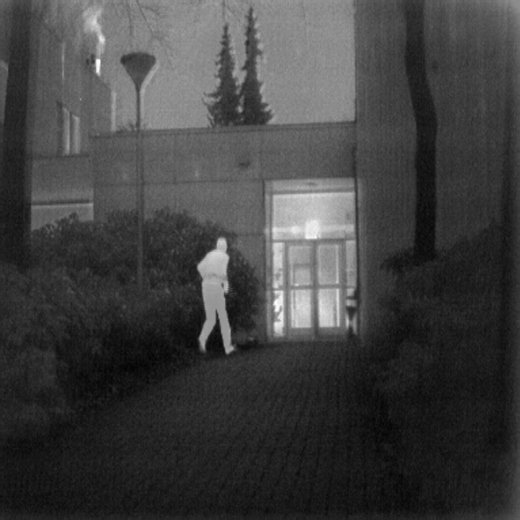}}
	{\includegraphics[height=0.12\linewidth]{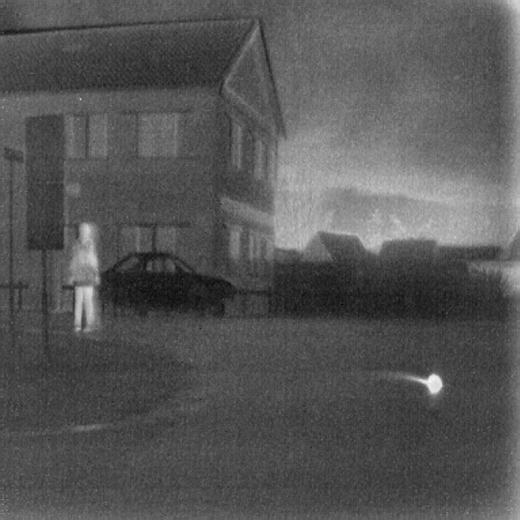}}
	{\includegraphics[height=0.12\linewidth]{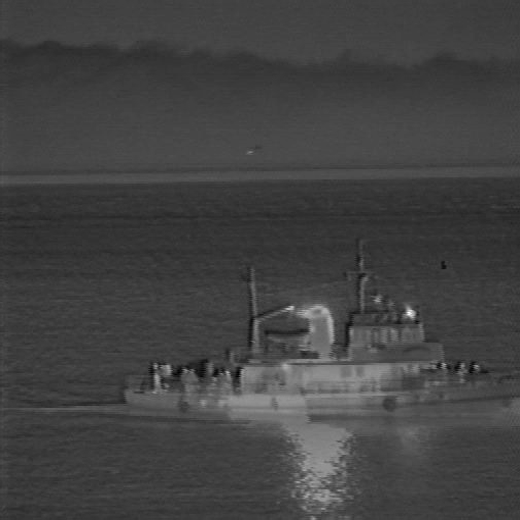}}
	{\includegraphics[height=0.12\linewidth]{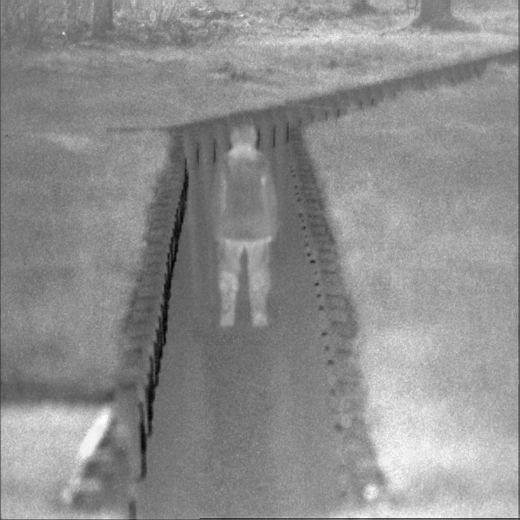}}
	{\includegraphics[height=0.12\linewidth]{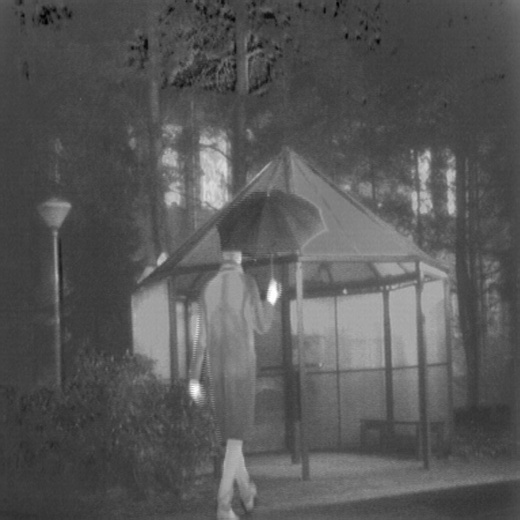}}
	\\
	\rotatebox{90}{GAN}
	{\includegraphics[height=0.12\linewidth]{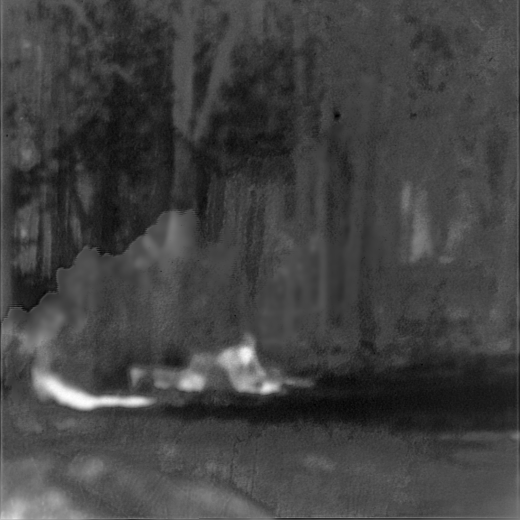}}
	{\includegraphics[height=0.12\linewidth]{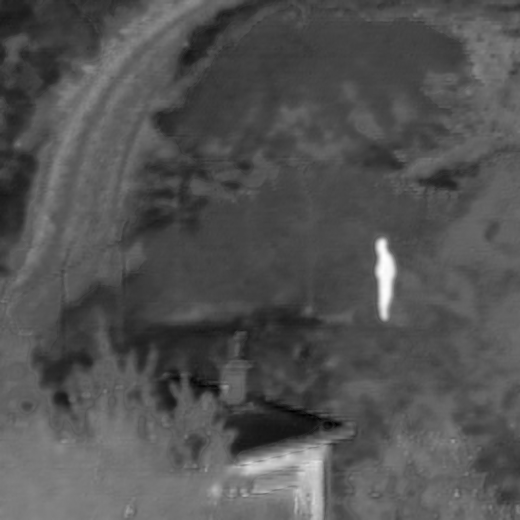}}
	{\includegraphics[height=0.12\linewidth]{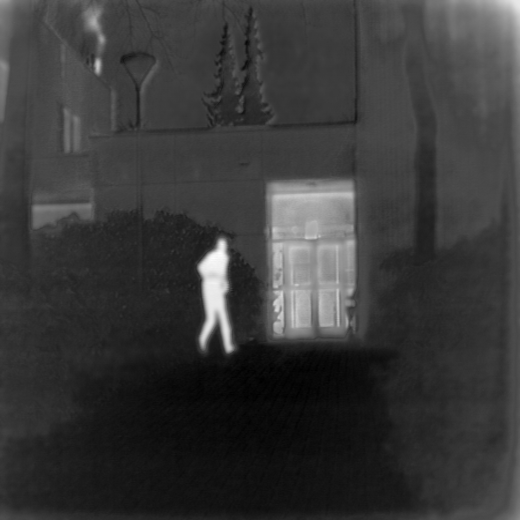}}
	{\includegraphics[height=0.12\linewidth]{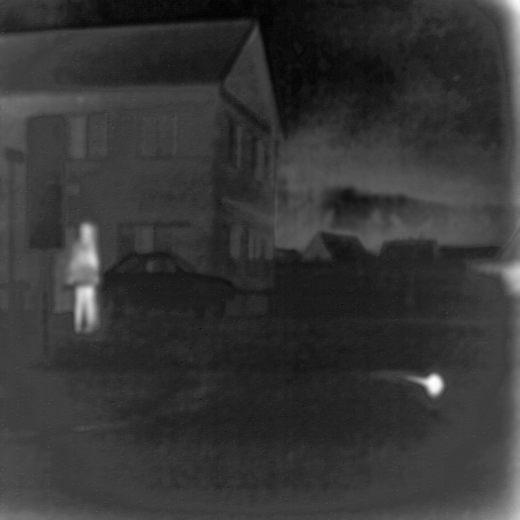}}
	{\includegraphics[height=0.12\linewidth]{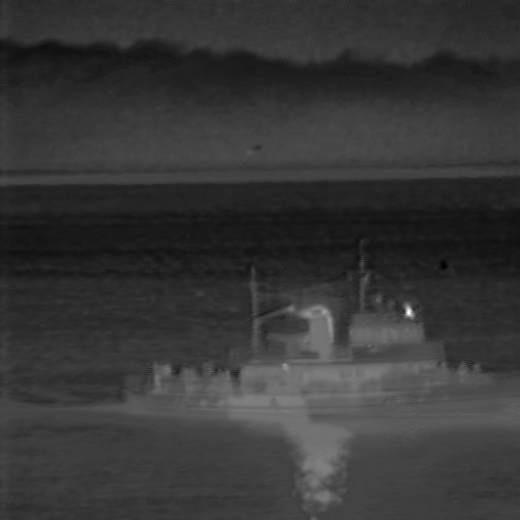}}
	{\includegraphics[height=0.12\linewidth]{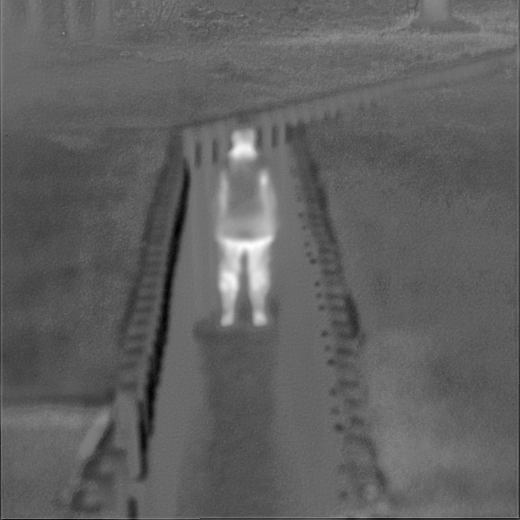}}
	{\includegraphics[height=0.12\linewidth]{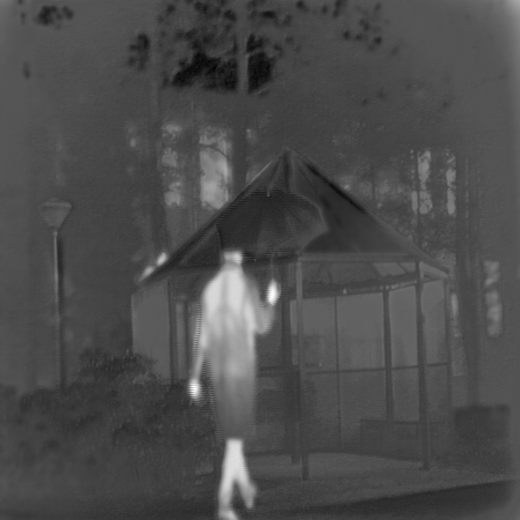}}
	\\
	\rotatebox{90}{Densefuse}
	{\includegraphics[height=0.12\linewidth]{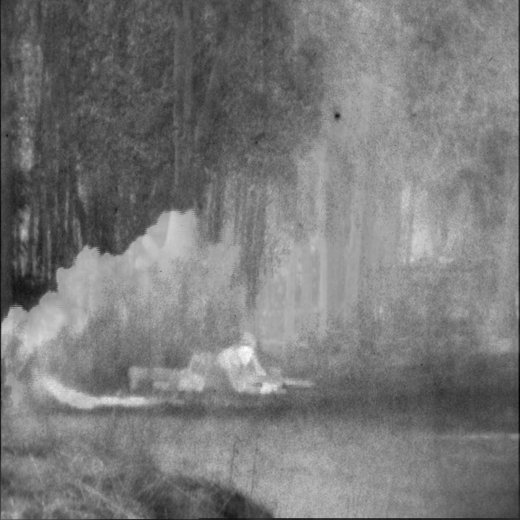}}
	{\includegraphics[height=0.12\linewidth]{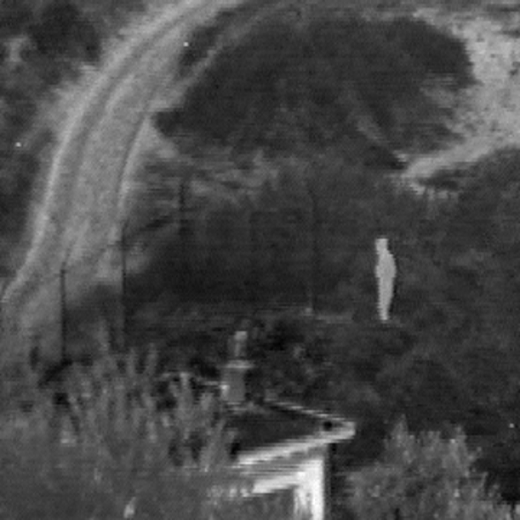}}
	{\includegraphics[height=0.12\linewidth]{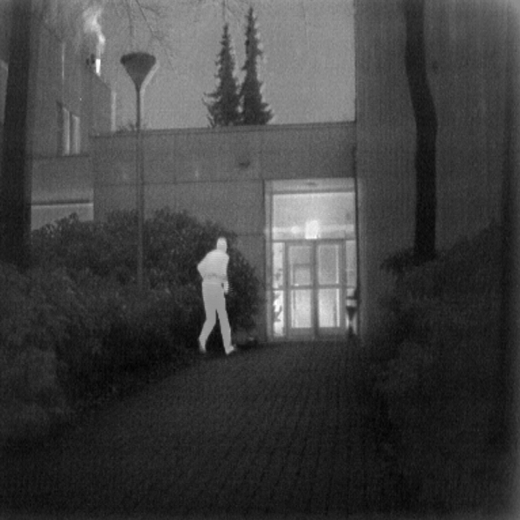}}
	{\includegraphics[height=0.12\linewidth]{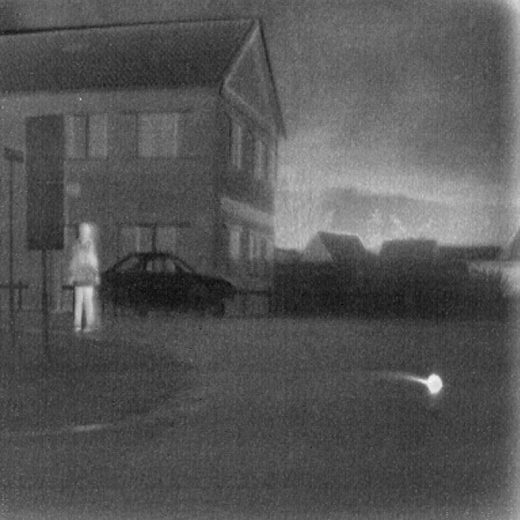}}
	{\includegraphics[height=0.12\linewidth]{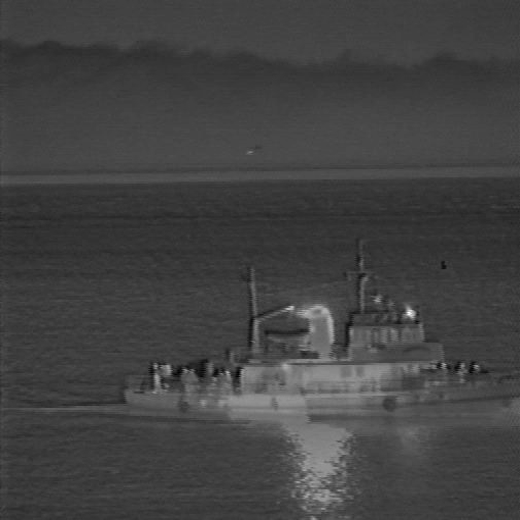}}
	{\includegraphics[height=0.12\linewidth]{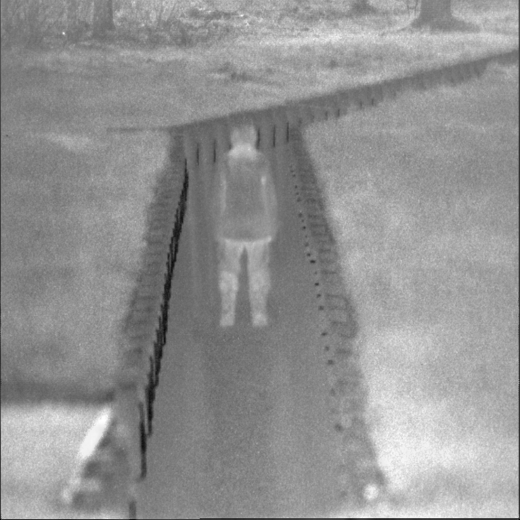}}
	{\includegraphics[height=0.12\linewidth]{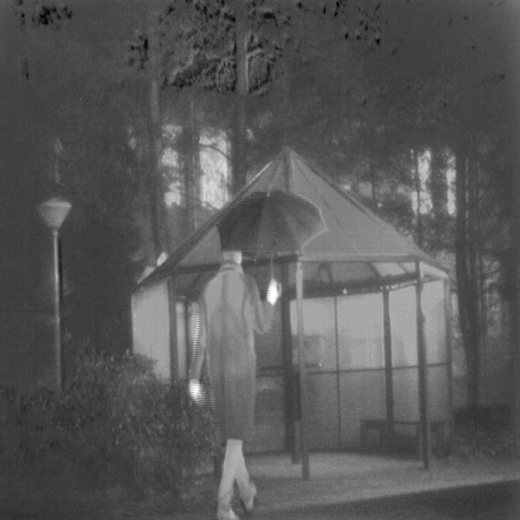}}
	\\
	\rotatebox{90}{IFCNN}
	{\includegraphics[height=0.12\linewidth]{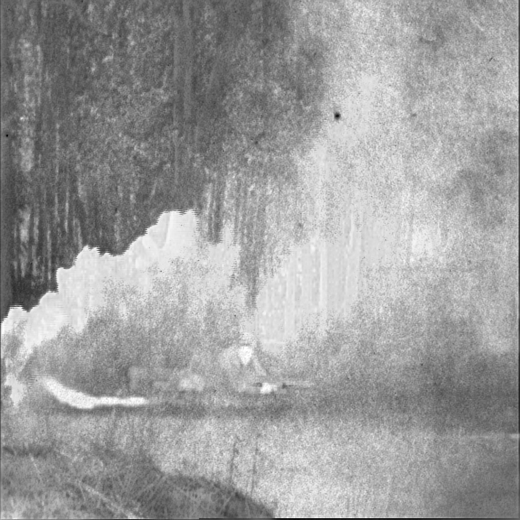}}
	{\includegraphics[height=0.12\linewidth]{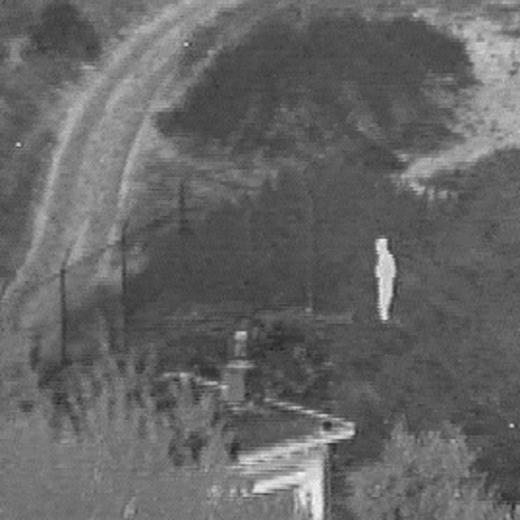}}
	{\includegraphics[height=0.12\linewidth]{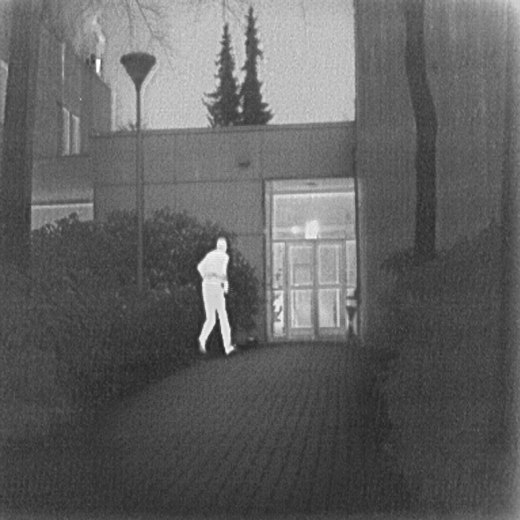}}
	{\includegraphics[height=0.12\linewidth]{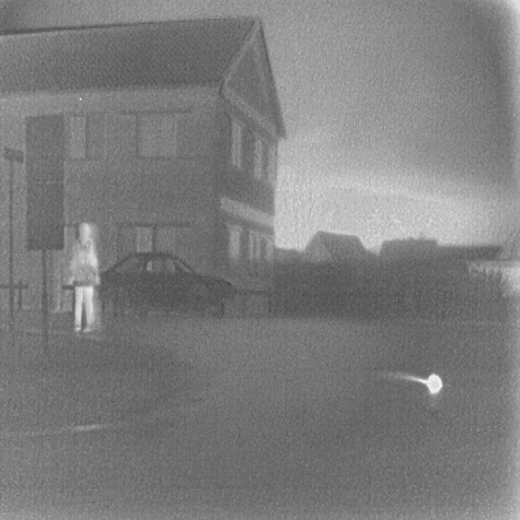}}
	{\includegraphics[height=0.12\linewidth]{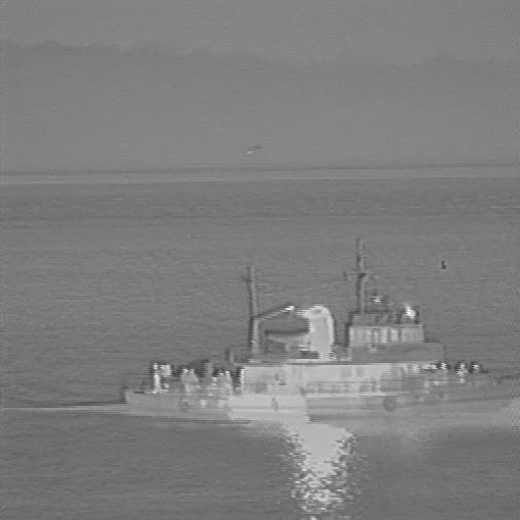}}
	{\includegraphics[height=0.12\linewidth]{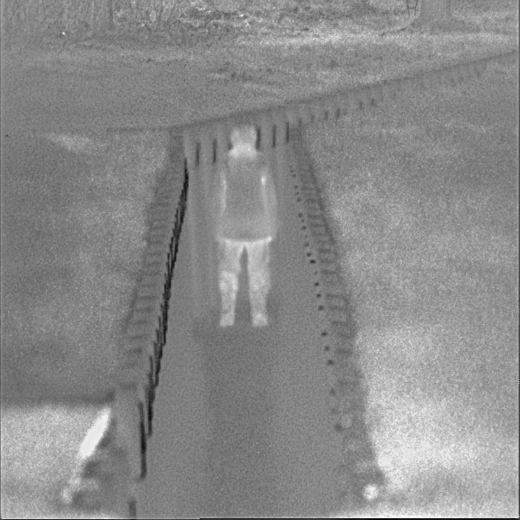}}
	{\includegraphics[height=0.12\linewidth]{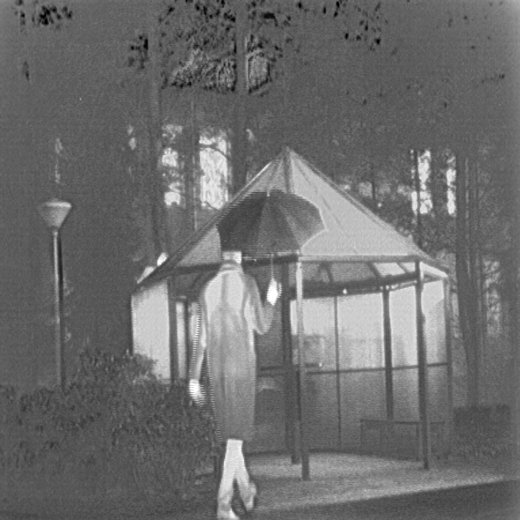}}
	\\
	\rotatebox{90}{Proposed}
	{\includegraphics[height=0.12\linewidth]{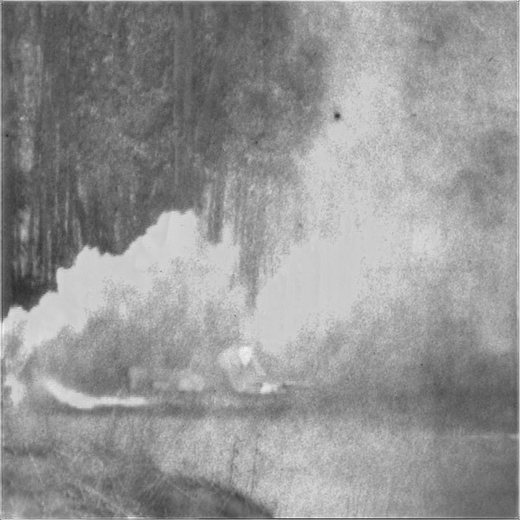}}
	{\includegraphics[height=0.12\linewidth]{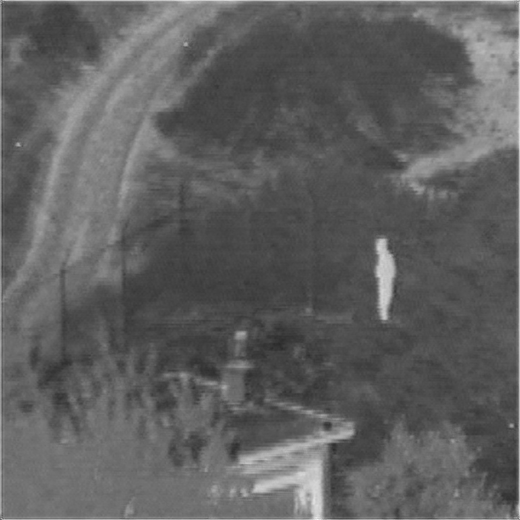}}
	{\includegraphics[height=0.12\linewidth]{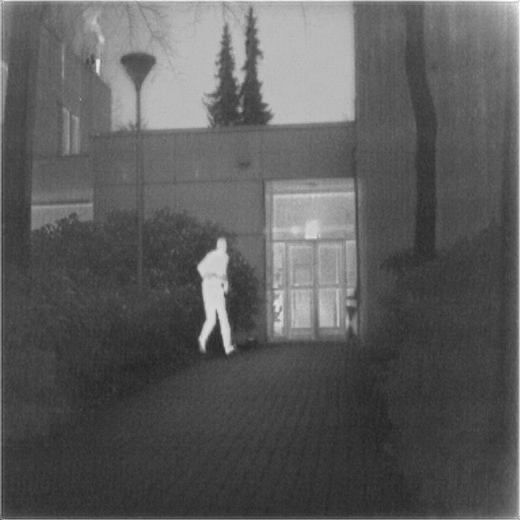}}
	{\includegraphics[height=0.12\linewidth]{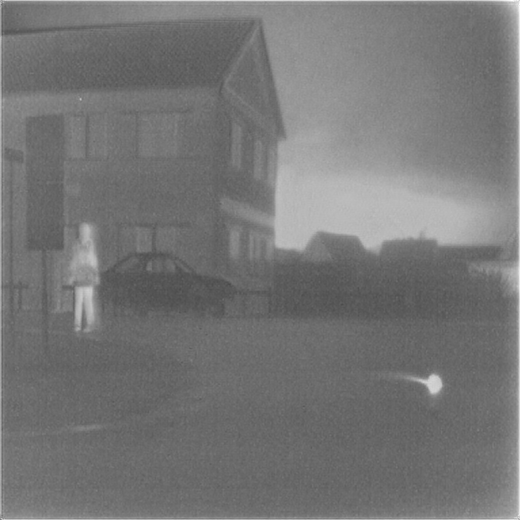}}
	{\includegraphics[height=0.12\linewidth]{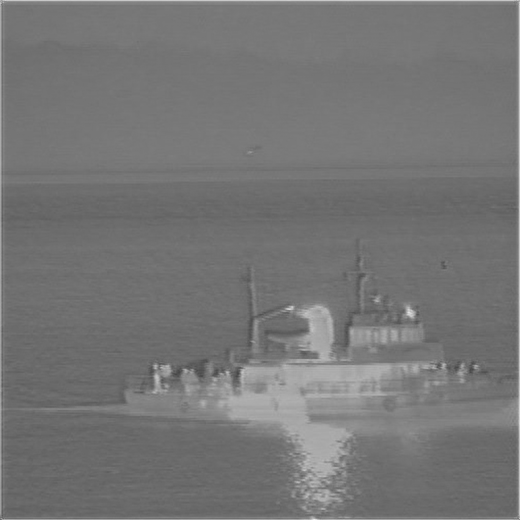}}
	{\includegraphics[height=0.12\linewidth]{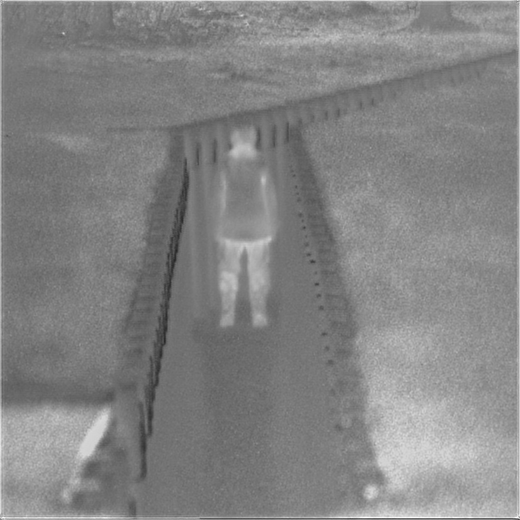}}
	{\includegraphics[height=0.12\linewidth]{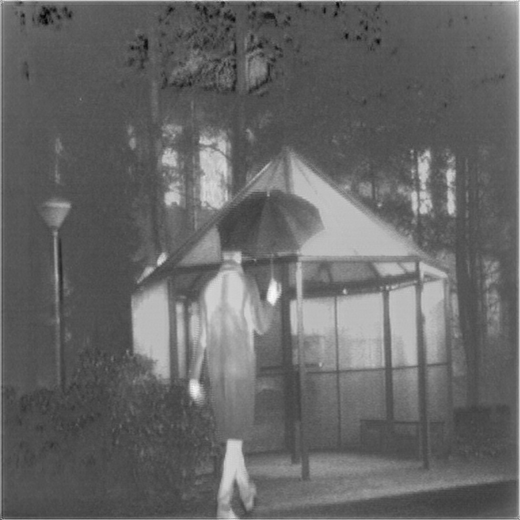}}
	\\
	\caption{From left to right(\dbname{Fog, Forest, Building, House, Boat, Solider, Umbrella}) are seven pairs of infrared and visible images to be fused, from up to down are the fusion results obtained by GFF, LPSR, GTF, CSR, MFCNN, Deepfuse, GAN, Densefuse, IFCNN and Proposed method.}
	\label{fig:undefine} 
\end{figure*}

In \autoref{tab:index}, the MFCNN method achieved the first place in the MI metric, but from the results of the fusion image, the MFCNN method integrates a pair of source images with significant pixel values into the fusion image. For example, in \dbname{``Fog''} in \autoref{fig:undefine}, the people in the smoke are directly discarded, which makes the fused image looks brighter and more complete, and the MI indicator is deceived during calculation. Our method has achieved the best results in most of the remaining deep learning methods. This benefits from our network framework and fusion rules. After extracting redundant and complementary information, corresponding fusion rules are designed according to their respective characteristics. So that the feature information is more complete, and the decoding layer has a better effect of regenerating this information into pictures. Compared with Densefuse, the image fusion method based on auto-encoding network, our improvements in feature classification and fusion rule retain information better. Significance indicators also show our differences. At the same time, we achieved the first place in the SSIM index. Adding the image quality function to the loss function allows our images to have better structure like \autoref{fig:multi-loss}. GAN generates images autonomously, with less artificial noise, so it performs well on CC indicators. From the significance index, we are no different from most deep learning methods except for the GAN method and better than traditional methods. In the same way, we are slightly worse than GAN on Q$_{cv}$, but better than most deep learning methods. Our fusion method retains the edges of the image better. Deepfuse, Densefuse and IFCNN directly add the pixel values when fusing the features, which makes the sum of the correlations of differences (SCD) performs better than us.

\section{Conclusion} \label{sec:conlusion}			
We have developed a novel deep learning framework named joint convolution auto-encoder network (JCAE) for infrared and visible image fusion. JCAE's advantaged structure with private and common branches can effectively learn the redundant and complementary relationships among infrared and visible paired images. To improve its feature learning ability, some layers of VGG19 were transferred into JCAE and a fusion metric relevant loss function was constructed. We have also proposed fusion rules respectively on the private branches and common branches to fuse deep feature maps into more information-rich maps, which then can be decoded and merged into a fused image by the JCAE. The subjective and quantitative evaluations on several metrics compared with other nine state-of-the-art fusion methods demonstrated that JCAE learned features and their corresponding fusion strategies benefit the infrared and visible image fusion.

\section{Compliance with Ethical Standards:}
This study was funded by the National Natural Science Foundation of P. R. China (grant number 61772237) and the Six Talent Peaks Project in Jiangsu Province (grant numbe XYDXX-030).

Ethical approval: This article does not contain any studies with human participants or animals performed by any of the authors.

%\clearpage
%\bibliographystyle{splncs04}
\bibliographystyle{unsrt} 
\bibliography{egbib}

\end{document}